\documentclass[fleqn,10pt]{wlscirep}
\usepackage[utf8]{inputenc}
\usepackage[T1]{fontenc}
\usepackage{longtable}
\usepackage{xcolor}
\usepackage{nameref}
\usepackage{hyperref}
\hypersetup{
    colorlinks=true,
    linkcolor=blue,
    filecolor=magenta,      
    urlcolor=cyan,
    pdftitle={Overleaf Example},
    pdfpagemode=FullScreen,
    }

\usepackage{longtable}
\usepackage{booktabs}
\usepackage{tabularx}
\usepackage{geometry}
\usepackage{makecell}
\usepackage{amsmath}
\usepackage{comment}
\usepackage{booktabs}
\usepackage{threeparttable}
\usepackage{adjustbox}
\usepackage{siunitx}

\newcolumntype{Y}{>{\raggedright\arraybackslash\hspace{0pt}\small}X}
\newcolumntype{S}{>{\centering\arraybackslash\hspace{0pt}\footnotesize}p{1.2cm}}
\newcolumntype{A}{>{\raggedright\arraybackslash\hspace{0pt}\tiny}p{3cm}}
\usepackage{multirow}   
\usepackage{booktabs} 
\newcommand*{\fullref}[1]{\hyperref[{#1}]{\autoref*{#1} \nameref*{#1}}}

\newcommand*{\subsecref}[1]{\hyperref[{#1}]{\nameref*{#1}}}

\title{Predicting acute food crises and discovering actionable drivers with causal machine learning}

\author[1,+,*]{Melissande Machefer}
\author[1,+]{Michele Ronco}
\author[2]{Michele Meroni}
\author[3]{Jordi Cerd\`a-Bautista}
\author[1]{Francesco Collivignarelli}
\author[5]{Chungmann Kim}
\author[6]{Nicolas Rost}
\author[7,8]{Alessia Matanó}
\author[3,4]{Esther Rodrigo-Bonet}
\author[9]{Philipp Zimmer}
\author[3]{Gustau Camps-Valls}
\author[1]{Christina Corbane}
\author[1]{Felix Rembold}

\affil[1]{European Commission, Joint Research Centre (JRC), Ispra, Italy}
\affil[2]{Seidor Consulting, Barcelona, Spain}
\affil[3]{University of Valencia, Valencia, Spain}
\affil[4]{now at Universidad Europea de Valencia, Valencia, Spain}
\affil[5]{Global Support Unit, Integrated Food Security Phase Classification (IPC), Rome, Italy}
\affil[6]{United Nations Office for the Coordination of Humanitarian Affairs (OCHA), New York, NY, and Harvard Humanitarian Initiative, Cambridge, MA, USA}
\affil[7]{Institute for Environmental Studies, Vrije Universiteit Amsterdam, Amsterdam, the Netherlands}
\affil[8]{now at Swedish Centre for Impacts of Climate Extremes (climes), Uppsala University, Uppsala, Sweden}
\affil[9]{World Bank Group, Washington, DC, USA}
\affil[*]{melissande.machefer@ec.europa.eu}
\affil[+]{these authors contributed equally to this work}

\begin{abstract}
Acute food insecurity is driven by a compound of conflict, adverse climate events, displacement, and economic shocks. We enrich the open-access Harmonized Food Insecurity Dataset with driver variables (monthly, subnational, 80 countries, 2007--2024) and develop missing-data-resilient machine learning pipelines forecasting crisis occurrence at 3, 6, and 9-month lead times using both FEWS NET and, for the first time at scale, IPC/CH classifications. Pipelines, evaluated through nested cross-validation, outperform no-skill baselines by 15 to 50 percentage points in PR-AUC, with performance correlated with crisis prevalence rather than the classification system. TreeSHAP-based explainability reveals compound, geography-specific predictive signal profiles, with three recurrent forecasting archetypes. Comparing predictive importance against doubly robust causal estimates of actionable drivers, we find that acute conflict events, having the weakest predictive signals, emerge as primary causal drivers of crisis classification whilst agroclimatology is the sole dimension in which predictive importance and causal effect align. 
\end{abstract}
\begin{document}

\flushbottom
\maketitle

\thispagestyle{empty}


\section*{Introduction}

In 2025, 266 million people across 47 countries, nearly one in five of the analyzed population, faced high levels of acute food insecurity (AFI)\cite{fsin2026grfc}. Conflict, economic shocks, and climate extremes, frequently co-occurring and mutually reinforcing\cite{fsin2026grfc, ronco2023exploring}, are the principal drivers, with impacts influenced by socioeconomic status, institutional resilience, and displacement\cite{anderson2021violent, nisbet2022food, cattaneo2026broadening}. These pressures are intensifying: conflicts are multiplying\cite{davies2025organized}, global displacement has reached record levels\cite{unhcr2025midyear}, and climate change is increasing the frequency and intensity of extreme weather events\cite{rentschler2022report, otto2023attribution}. Famines declared multiple times in Somalia, South Sudan, Sudan, and Gaza within the past fifteen years were ultimately products of political decisions — conflict, obstruction of aid, and governance failure — not environmental inevitability\cite{dewaal2018end}.

Tracking these dynamics requires timely, granular early warning that reflects the compound, multidimensional nature of the drivers themselves \cite{richards2021, ronco2026}. Two compatible classification systems provide essential global monitoring: the Integrated Food Security Phase Classification/Cadre Harmonisé (IPC/CH) and the Famine Early Warning Systems Network (FEWS NET). Both use a comparable five-phase scale classification — from none or minimal food insecurity to famine — and deliver subnational assessments alongside short-term projections three to six months ahead. Yet, their methodologies differ: IPC/CH relies on a consensus-based convergence-of-evidence process, while FEWS NET draws on expert-driven scenario analysis\cite{michelson2026integrated,lentz2026black, lentz2025official, machefer2025monthly}. Global coverage remains uneven, with each system monitoring fewer than 60 countries with inconsistent frequency, limiting comparability across contexts and leaving critical gaps in early warning.

Machine learning (ML) offers a promising path toward more cost-effective, timely, and scalable food insecurity estimations\cite{machefer2025potential, martini2022machine}. Recent studies span nowcasting and forecasting up to twelve months ahead at subnational level, leveraging conflict data, satellite-derived agroclimatic indicators, market prices, and food consumption surveys across multiple countries\cite{machefer2025potential, penson2024datadriven, van2024unsupervised}. However, generalizability remains an ongoing challenge. Most studies are limited in geographic scope, covering only a handful to at most twenty countries, and focus exclusively on modeling the FEWS NET phases\cite{machefer2025potential, van2024unsupervised}; IPC/CH phases, despite their centrality in humanitarian decision-making, have never been modeled at scale\cite{penson2024datadriven}. Furthermore, missing input data, including previous assessments, are rarely addressed, which limits these models where data are absent, delayed, or contested\cite{wang2022transitions, westerveld2021forecasting, lentz2019data, busker2024predicting, herteux2024forecasting}.

These limitations matter. Early warning linked to financing can trigger anticipatory action before crises escalate\cite{premand2022cash, dunsch2025impact, baylis2025impact}, yet funding for anticipatory action frameworks was only 1\% of international humanitarian assistance in 2025 \cite{alnap2026global}, and global humanitarian budgets are now contracting sharply\cite{osendarp2025lethal,locks2025shock,rega2025impact, hayward2026aid, cavalcanti2025usaid, rohner2026aiding}. Two further gaps remain underaddressed. Existing ML models are rarely validated across both major classification systems simultaneously, leaving it unclear whether predictive skill is system-specific or genuinely transferable. And while explainability tools such as SHapley Additive exPlanations (SHAP)\cite{lundberg2017unified} have made model predictions increasingly interpretable, translating feature importance into actionable prevention design requires causal evidence, something the food security literature has only begun to address\cite{tarraga2024causaldiscovery, cerda24causal, rodrigobonet25hcg}. Interpretability alone is also insufficient for responsible deployment\cite{campsvalls2025}. Ethical early warning systems need continuous oversight, reproducibility, expert input, and risk mapping \cite{anderson2026, OCHA2023, kreutzer2025ethical, reichstein2025}; the debate over Google’s flood model \cite{li2026ai, nearing_2026_19350220} shows the cost of lacking them, especially in food security, where lives and billions in aid are at stake.

We address these gaps with a three-stage ML framework combining forecasting, explainability, and causal inference. By extending the Harmonized Food Insecurity Dataset (HFID) \cite{machefer2025monthly} with predictors across six dimensions (agroclimatology, conflict, economics, food security, disaster risk, and population exposure), we build the largest open-access food insecurity modeling dataset to date: monthly, subnational, and covering 80 countries over the period 2007--2024, with fully open code and data. In the first stage, we benchmark multiple ML pipelines — including missing-data-resilient ensemble and transformer models such as Histogram-based gradient boosting (HistXGBoost)\cite{guryanov2019, hoang2023} and the Tabular Prior-data Fitted Network (TabPFN)\cite{hollmann2025} — to forecast monthly subnational AFI targets (crisis occurrence, emergency occurrence, crisis escalation) at three-, six-, and nine-month lead times under both IPC/CH and FEWS NET. This is the first study to model both systems in parallel at a global scale, evaluated through a rigorous nested cross-validation strategy combining multiple metrics, naive baseline comparisons, and statistical significance testing. In the second stage, we apply TreeSHAP-based explainability to a performance-weighted combination of both classification models, characterizing predictive signal profiles at the level of individual features and aggregated thematic dimensions, and tracking their spatial and temporal variation across countries and macro-regions. In the third stage, we estimate the Average Treatment Effect (ATE) of acute shocks on crisis probability and directly compare these causal estimates against SHAP-based feature importances. This disentangles what the model relies on for forecasting from the root causes relevant to structural intervention, a distinction with direct implications for intervention design, targeted policies and resource allocation.

\section*{Results}
\subsection*{Robust aggregate performance across lead times and classification systems} 

We evaluate three targets: crisis occurrence (IPC phase $\geq 3$), emergency occurrence (IPC phase $\geq 4$), and crisis escalation (i.e., a deterioration in food security resulting in a transition to IPC Phase 3  or worse). In a first experiment, historical target values are excluded from the feature set to reflect operational conditions when they are unavailable. Only crisis occurrence performs satisfactorily, with PR-AUC and F1 scores exceeding 0.5 across pipelines (Supplementary Fig. \ref{fig:perf_radar}); the other two targets suffer from severe class imbalance, with positive class prevalence below 5\% in half of the test folds (Supplementary Table \ref{tab:cv_summary}). Subsequent analyses are therefore focused on crisis occurrence.

\begin{figure}[h!]
\includegraphics[width=0.95\textwidth]{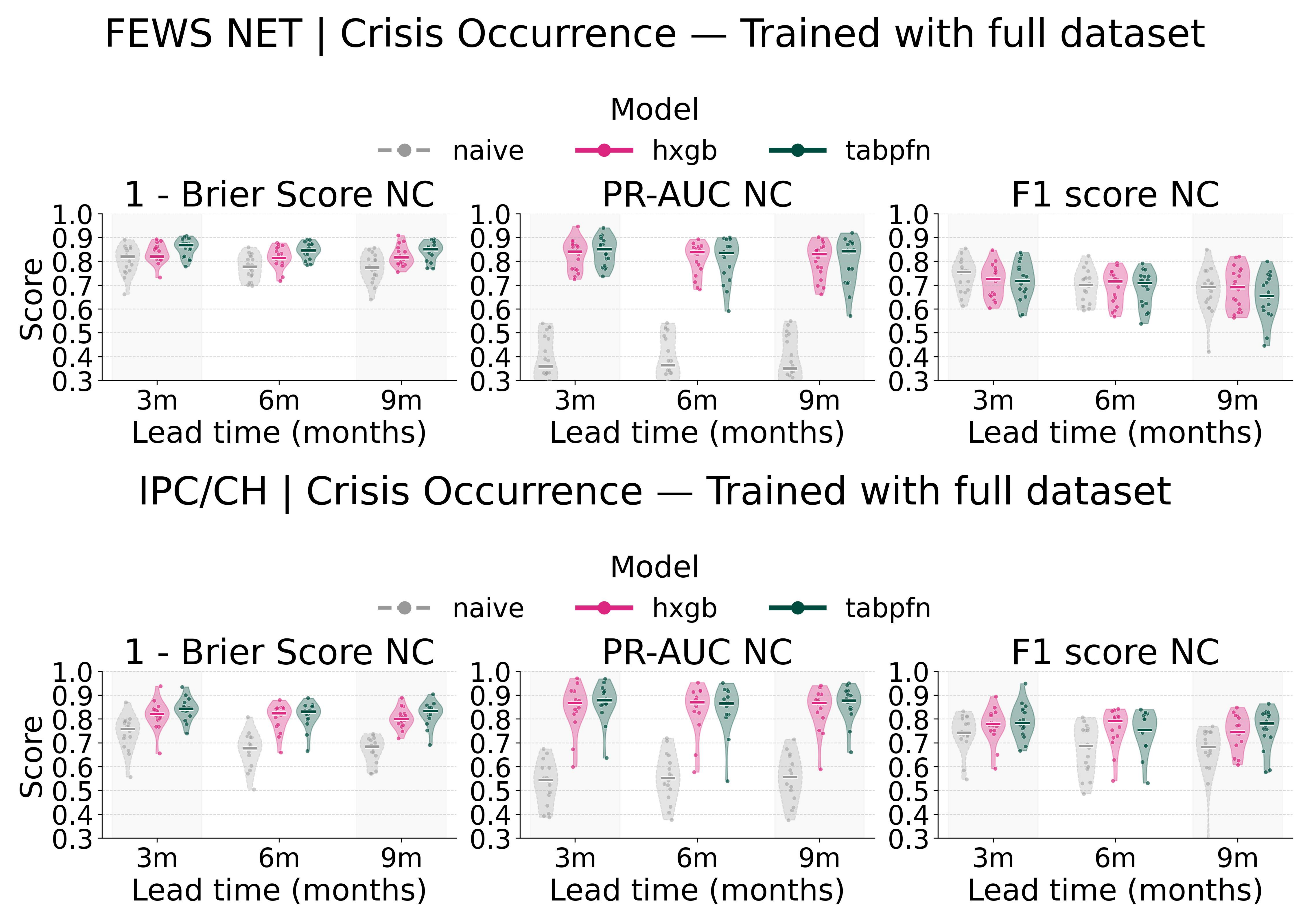}
\caption{Cross-validated model performance for crisis occurrence forecasting. Outlying points, violin distributions, and median horizontal bars depict per-fold test metrics for folds with a positive rate $>10\%$, restricted to non-constant (NC) administrative areas. HistXGBoost (hxgb) and TabPFN (tabpfn) handle missing values and are trained/evaluated on the ``Full dataset''. The complement of the Brier Score is shown ($1 - \text{Brier Score}$), so that higher values indicate superior performance across all metrics.}
\label{fig:perf_violin_full}
\end{figure}

Ensemble and foundation models (Random Forest, XGBoost, HistXGBoost, and TabPFN\cite{hollmann2025}) consistently outperform Logistic Regression and Decision Tree baselines across all metrics and both classification systems (Supplementary Fig. \ref{fig:perf_radar}). With the lagged target included in the features, missing-data-resilient models (HistXGBoost, TabPFN) trained on the ``Full dataset'' (including missing data) perform comparably to  pipelines requiring complete data from the ``Restricted'' dataset (Fig. \ref{fig:perf_violin_full} and Supplementary Fig. \ref{fig:perf_violin_restricted}).  All models outperform the naive persistence baseline and exceed no-skill PR-AUC by 15 to 50 percentage points under IPC/CH. Under FEWS NET, F1 gains over the baseline are smaller, driven by a lower positive-class prevalence (median $\approx 0.35$ versus $\approx 0.50$ for IPC/CH), while probabilistic discrimination remains robust. Removing the lagged target as a feature lowers  F1 score by up to 20 percentage points, but by no more than 7 points across all metrics on the ``Full dataset'' (Supplementary Fig. \ref{fig:includeprevious}),indicating that driver features alone carry sufficient predictive signal for missing-data resilient pipelines. Given their strong performance and ability to handle sparse data, we restrict subsequent analyses to missing-data-resilient models (TabPFN and HistXGBoost). Their skill remains stable across lead times (Fig. \ref{fig:perf_violin_full} and Supplementary Fig. \ref{fig:perf_violin_restricted}), so we report results at the three-month horizon in the following.

\begin{table}[h!]
\centering
\caption{Pairwise Wilcoxon signed-rank $p$-values and Nemenyi post-hoc $p$-values 
on Brier Score. Significant results ($\alpha=0.05$) are marked with an asterisk (*). Mean ranks 
are derived from the Friedman test (lower is better). Significance tests are carried out across outer folds for TabPFN (tabpfn), 
HistXGBoost (hxgb) and the naive baseline, for crisis occurrence at 
3-month lead time. Results are reported on the ``Full dataset'' (All) and for non-constant administrative areas (NC).}
\label{tab:appendix_sig}
\resizebox{\textwidth}{!}{%
\begin{tabular}{ll cc cc}
\toprule
& & \multicolumn{2}{c}{\textbf{All}} & \multicolumn{2}{c}{\textbf{Non-Constant (NC)}} \\
\cmidrule(lr){3-4} \cmidrule(lr){5-6}
\textbf{Source} & \textbf{Comparison} 
  & \textbf{Wilcoxon $p$} & \textbf{Nemenyi $p$}
  & \textbf{Wilcoxon $p$} & \textbf{Nemenyi $p$} \\
\midrule
\multirow{3}{*}{IPC/CH}
  & tabpfn vs hxgb  & $.083$     & $.539$     & $.004^{*}$ & $.539$     \\
  & tabpfn vs naive & $.002^{*}$ & $.001^{*}$ & $.000^{*}$ & $.000^{*}$ \\
  & hxgb vs naive   & $.003^{*}$ & $.022^{*}$ & $.000^{*}$ & $.001^{*}$ \\
\cmidrule(l){2-6}
  & \multicolumn{3}{l}{Mean ranks (All): tabpfn $= 1.44$, hxgb $= 1.81$, naive $= 2.75$} 
  & \multicolumn{2}{l}{Mean ranks (NC): tabpfn $= 1.31$, hxgb $= 1.69$, naive $= 3.00$} \\
\midrule
\multirow{3}{*}{FEWS NET}
  & tabpfn vs hxgb  & $.001^{*}$ & $.051$     & $.009^{*}$ & $.112$     \\
  & tabpfn vs naive & $.000^{*}$ & $.001^{*}$ & $.000^{*}$ & $.000^{*}$ \\
  & hxgb vs naive   & $.495$     & $.377$     & $.018^{*}$ & $.033^{*}$ \\
\cmidrule(l){2-6}
  & \multicolumn{3}{l}{Mean ranks (All): tabpfn $= 1.33$, hxgb $= 2.11$, naive $= 2.56$} 
  & \multicolumn{2}{l}{Mean ranks (NC): tabpfn $= 1.28$, hxgb $= 1.94$, naive $= 2.78$} \\
\bottomrule
\end{tabular}}
\end{table}
TabPFN performs best on non-constant administrative units (that experience temporal target variation), with a PR-AUC of $0.87 \pm 0.08$ and a Brier Score of $0.16 \pm 0.05$ for IPC/CH, and $0.84 \pm 0.06$ and $0.14 \pm 0.04$ for FEWS NET (Supplementary Table \ref{tab:appendix_metrics}). TabPFN and HistXGBoost both significantly outperform the naive baseline on Brier Score for IPC/CH; for FEWS NET, HistXGBoost's advantage is significant only in non-constant areas experiencing genuine transitions (Table \ref{tab:appendix_sig}). We use HistXGBoost for downstream analysis because it supports exact TreeSHAP decomposition (see Methods), and its performance remains stable across outer folds (Supplementary Fig. \ref{fig:temp_fold}), with a modest positive trend with increasing sample size, indicating that reported averages are conservative estimates of operational skill.

Country-level performance correlates strongly with positive class prevalence (Spearman correlation with PR-AUC exceeding 0.92 for IPC/CH and 0.56 for FEWS NET; Supplementary Figs. \ref{fig:perf_country_fews}, \ref{fig:perf_tab_country_fews},  \ref{fig:perf_country_ipcch} and \ref{fig:perf_tab_country_ipcch}). High-prevalence countries such as Yemen, South Sudan, Haiti, and the Central African Republic achieve near-perfect discrimination, while low-prevalence settings (below 5\%, such as Burundi and Mali) exhibit weak PR-AUC regardless of data volume: Nigeria contributes over 11,000 test observations under IPC/CH yet achieves a PR-AUC of only 0.429. Pooled scores are similar across systems (PR-AUC and F1 of 0.885 and 0.829 for IPC/CH; 0.876 and 0.792 for FEWS NET). Afghanistan's PR-AUC gap between IPC/CH (0.963) and FEWS NET (0.404) likewise reflects differences in positive class prevalence (IPC/CH 90.7\% versus FEWS NET 29.3\%), not a better fit to either system: IPC/CH coverage starts only in 2017, a period of frequent crises in the country, compared to previous years (FEWS NET starts in 2011).

\begin{figure}[h!]
\centering
\includegraphics[width=18.5cm,height=11.5cm]{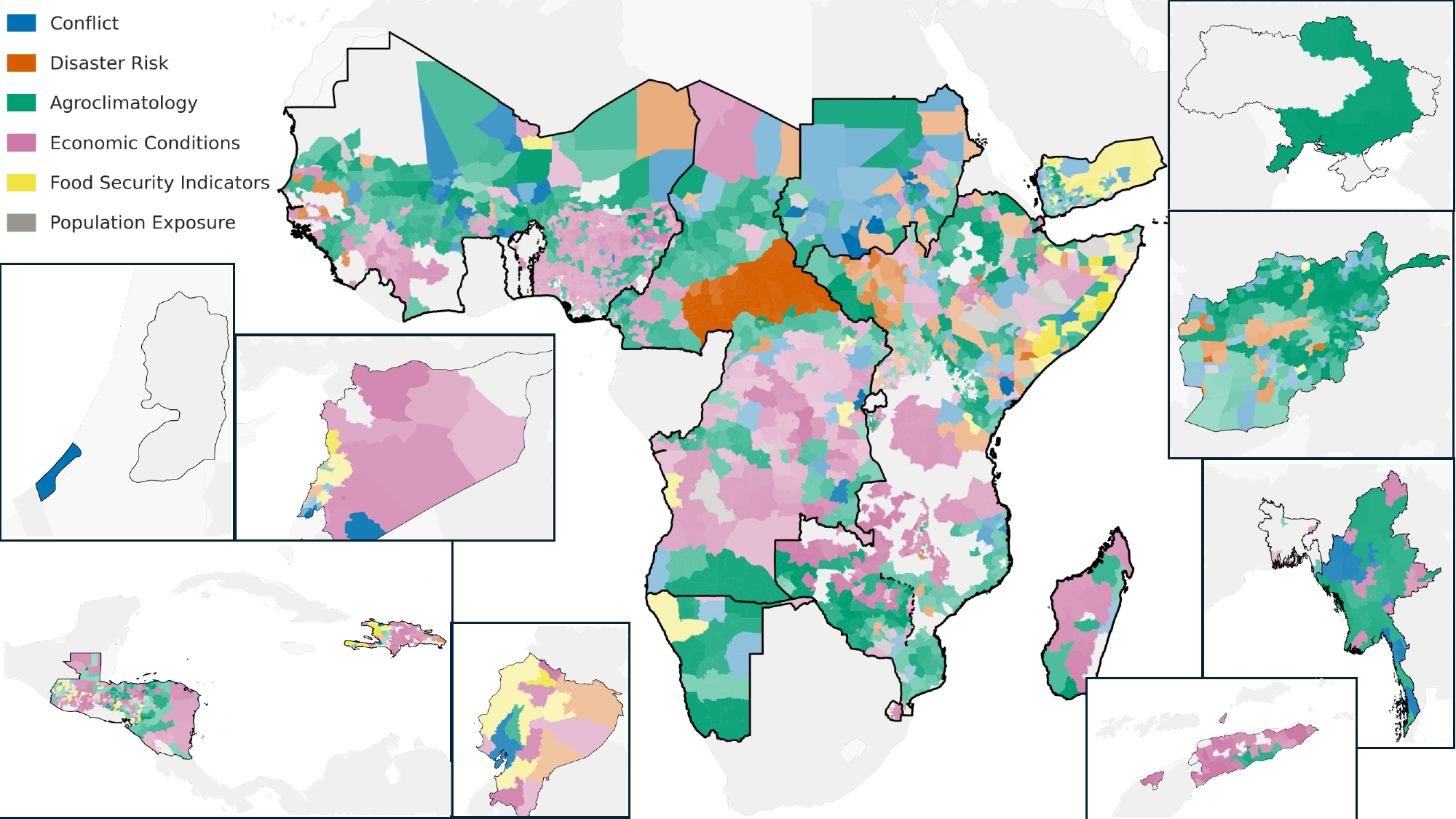}
\caption{Primary emerging predictive signals of food insecurity. Colors indicate the dominant risk dimension for each Admin-2 region over the 18-month model test period (Jan 2023–May 2024), synthesized from performance-weighted predictive models for both IPC/CH and FEWS NET crisis levels. The designated dimension identifies the thematic domain that combined the highest absolute attribution weight with the strongest growth over the timeline. Color saturation indicates the intensity of this trend, highlighting regions where risk features accelerate. Bold outlines denote macro-regional boundaries.}
\label{fig:shap_map}
\end{figure}

\subsection*{Explainability reveals compound and regionally heterogeneous predictive signals} 

To characterize the predictive profiles of acute food crises across both classification systems and geographies, we apply TreeSHAP \cite{lundberg2017unified} to two HistXGBoost classifiers, one optimized for FEWS NET and one for IPC/CH crisis levels. Both were trained on records before January 2023 and tested on held-out data from January 2023 to May 2024 across 50 countries (FEWS NET: F1 = 0.76, PR-AUC = 0.72; IPC/CH: F1 = 0.79, PR-AUC = 0.80). We combine their attributions into a performance-weighted unified signal over 74,897 subnational ($\text{Admin-2} \times \text{month}$) test predictions where at least one model predicted a crisis (see Methods). Feature importances are then aggregated into six thematic dimensions: Agroclimatology, Conflict, Economic Conditions, Food Security Indicators, Disaster Risk, and Population Exposure, following Machefer et al. \cite{machefer2025potential} (see Table \ref{tab:feature_categories}).

\begin{figure}[h!]
\centering
\includegraphics[width=18.5cm,height=12cm]{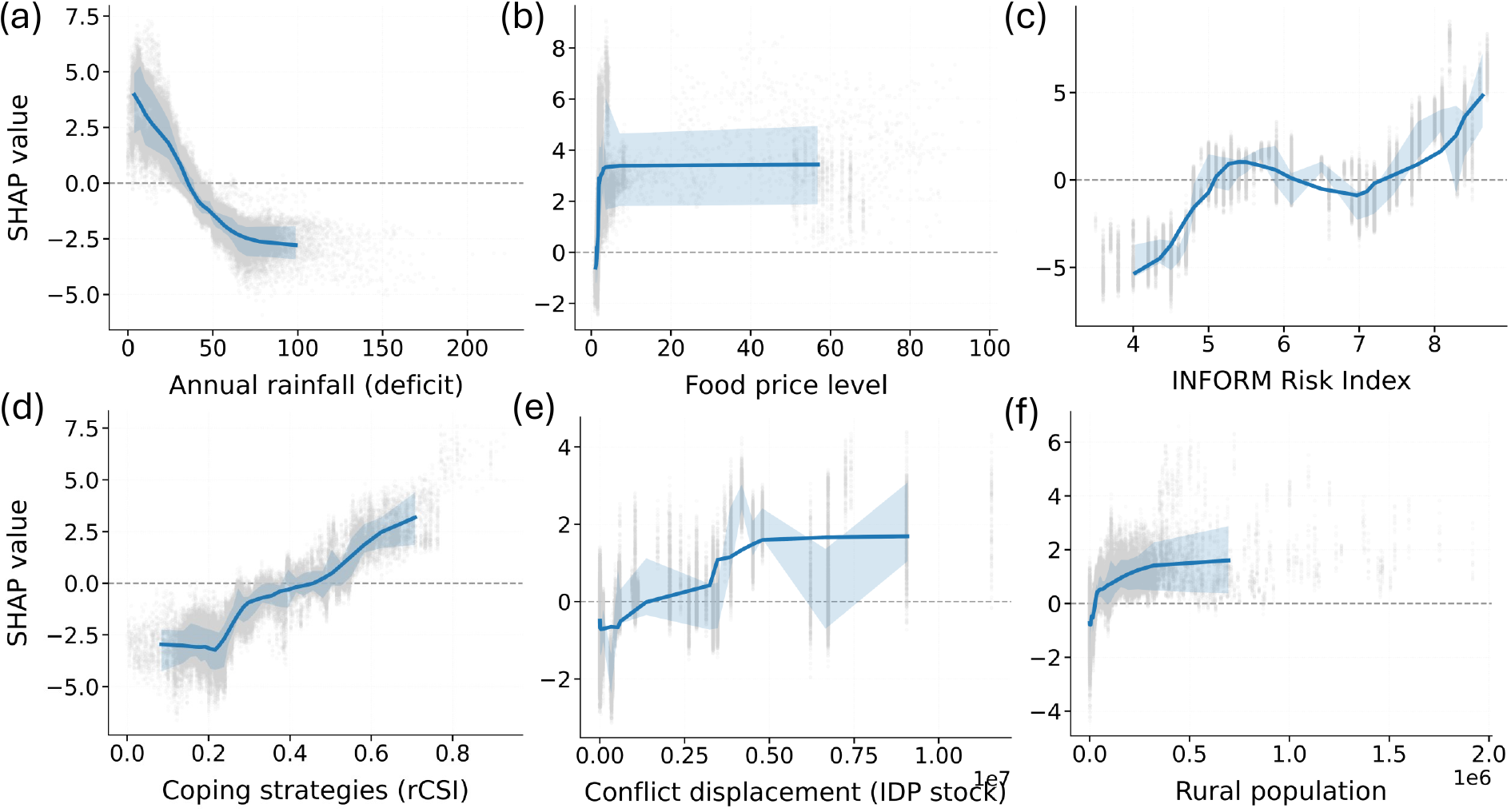}
\caption{Nonlinear driver responses. Dependence plots illustrate the marginal effects of representative predictors from six semantic feature groups on the probability of a food crisis. For each group, the single most influential input variable was selected based on mean absolute SHAP values. The vertical axis shows the smoothed mean SHAP response, where positive values indicate increased crisis risk. Panels show: (a) Annual rainfall (deficit); (b) Food price level; (c) INFORM Risk Index; (d) Coping strategies (rCSI)); (e) Conflict displacement (IDP stock); and (f) Rural population. Solid lines represent smoothed mean responses, while shaded areas indicate ±1 standard deviation across quantile bins.}
\label{fig:scatshap}
\end{figure}

Globally, no single thematic dimension dominates; the leading predictive signal varies by region, mirroring the heterogeneous and compound dynamics of contemporary food crises \cite{fsin2023grfc, fsin2024grfc} (Fig.~\ref{fig:shap_map}). Yet, when aggregated across regions, Economic Conditions and Agroclimatology emerge as the two anchor dimensions, together accounting for approximately $57\%$ of total mean absolute SHAP importances (Supplementary Figs. \ref{fig:shap_summary}, \ref{fig:shap_summary_fews}, \ref{fig:shap_summary_ipcch}, \ref{fig:global_shap_bar}).  Within Agroclimatology, annual rainfall and FPAR carry the strongest predictive weight. Importance for yearly weighted-average dekadal rainfall turns markedly positive below approximately 35 mm per dekad (Fig.~\ref{fig:scatshap}a),  the threshold below which the ensemble relies heavily on precipitation  to forecast crisis. Within the Economic domain, the unified SHAP importance rises sharply once food prices exceed 1.5 standard deviations above the seasonal baseline (Fig.~\ref{fig:scatshap}b), a threshold shared by both classification systems that marks severe market disruption as a strong discriminator of crisis  \cite{andree2020predicting, andree2023machine, fsin2023grfc, fsin2024grfc}. Conflict and Disaster Risk each account for roughly $13\%$ of global attribution (Supplementary Fig. \ref{fig:global_shap_bar}). Disaster Risk is dominated by the INFORM Risk Index, which increases sharply above 7.3 out of 10 (Fig.~\ref{fig:scatshap}c), matching the framework's ``very high'' risk tier\cite{marzi2021assessing}.  Within Conflict, the cumulative stock of internally displaced persons carries the most weight, spiking sharply above 1.4 million at national level(Fig.~\ref{fig:scatshap}e). The remaining share is split between Food Security Indicators and Population Exposure. The rCSI turns positive at a $46\%$ prevalence (Fig.~\ref{fig:scatshap}d), where widespread coping behavior signals a likely crisis three months later\cite{martini2022machine}. Rural population  adds a modest positive weight above 27,000 inhabitants per Admin-2 unit (Fig.~\ref{fig:scatshap}f), suggesting that risk concentrates in larger agrarian-dependent populations.

\begin{figure}[h!]
\centering
\includegraphics[width=18.5cm,height=7.5cm]{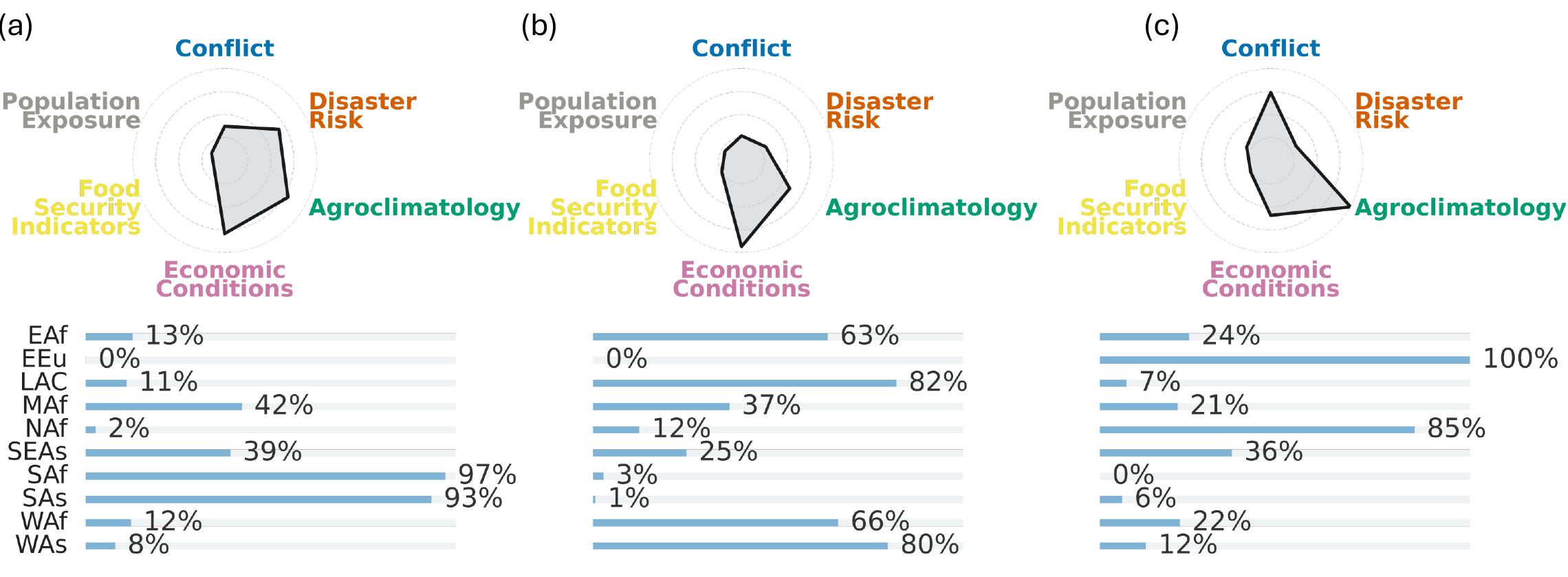}
\caption{Cluster-based relative predictive profiles and regional composition of food crisis forecasts. Panels a, b, and c refer to archetypes 0, 1, and 2, respectively. Radar plots display the mean unified SHAP importance for key predictive categories across three distinct clusters obtained via row-wise min-max scaling and direct $K$-means clustering. Horizontal bars show the percentage distribution of subnational units (Admin-2) belonging to each forecasting archetype across ten macro-regions: Eastern Africa (EAf), Eastern Europe (EEu), Latin America and the Caribbean (LAC), Northern Africa (NAf), Middle Africa (MAf), South-Eastern Asia (SEAs), Southern Africa (SAf), Southern Asia (SAs), Western Africa (WAf), and Western Asia (WAs).}
\label{fig:clust}
\end{figure}

Macro-regional and subnational signal profiles reveal systematic geographic gradients alongside sharp temporal shifts (see Fig.~\ref{fig:shap_map}, and Supplementary Fig.~\ref{fig:shap_temporal}). Clustering $\text{Admin-2}$ units by relative unified SHAP importance yields three recurrent forecasting archetypes, demonstrating that the ensemble relies on compound signals rather than single indicators (Fig.~\ref{fig:clust}). Archetype 0 (agroclimatic-structural vulnerability) combines Agroclimatology ($\sim 27\%$), Economic Conditions ($\sim 27\%$), and Disaster Risk ($\sim 23\%$) across Southern Africa (SAf), Southern Asia (SAs), and Middle Africa (MAf), where climate exposure intersect  structural baseline fragility. Archetype 1 (market-dependency) is led by Economic Conditions ($\sim 36\%$) and Agroclimatology ($\sim 24\%$), and covers most units in  Western Africa (WAf), Eastern Africa (EAf), Latin America and the Caribbean (LAC), and Western Asia (WAs). Archetype 2 (conflict-agricultural) pairs Conflict ($\sim 23\%$) with Agroclimatology ($\sim 31\%$), characterizing Northern Africa (NAf), Eastern Europe (EEu), and specific pockets of WAs where displacement, conflict occurrence and agricultural disruption co-occur. Between January 2023 and April 2024, economic and agroclimatic contributions were generally more stable than conflict- and disaster-related contributions, though all dimensions varied over time, especially in 2024 (Supplementary Fig.~\ref{fig:shap_temporal}).

Geographically, the archetypes map onto distinct regional footprints. In Archetype 0, Agroclimatology dominates in SAs ($73\%$ of units) and SAf ($48\%$), capturing severe El Niño-driven drought in countries such as Zambia in early 2024 \cite{fsin2024grfc, fsin2025grfc} (Supplementary Fig.~\ref{fig:country_trajectories2}b). In Archetype 1, Economic Conditions dominate WAf ($68\%$ of units), reflecting pervasive price inflation across the Sahel and Gulf of Guinea, and in LAC ($47\%$), where import-price vulnerability is high, as illustrated by rising food price prices in Haiti\cite{fsin2024grfc} (Supplementary Fig.~\ref{fig:country_trajectories2}c). Archetype 2 is more heterogeneous. Conflicts dominate in active combat arenas such as Sudan ($45\%$ of units\cite{fsin2024grfc, Mohamed2025}) and Palestine, where the  profile shifts abruptly in October 2023 with a $79\%$ month-on-month acceleration in conflict SHAP values (Supplementary Fig.~\ref{fig:country_trajectories1}a). Conversely, in Ukraine (EEu), Agroclimatology exhibits the largest absolute unified SHAP importance (Fig.~\ref{fig:shap_map}), consistent with wartime damage to agricultural land\cite{rosa2024ukraine, snizhko2024ukraine}, followed by a sharp acceleration in the economic attribution from January 2024 (Supplementary Fig.~\ref{fig:country_trajectories1}d).

\begin{figure}[h!]
\centering
\includegraphics[width=18cm,height=18cm]{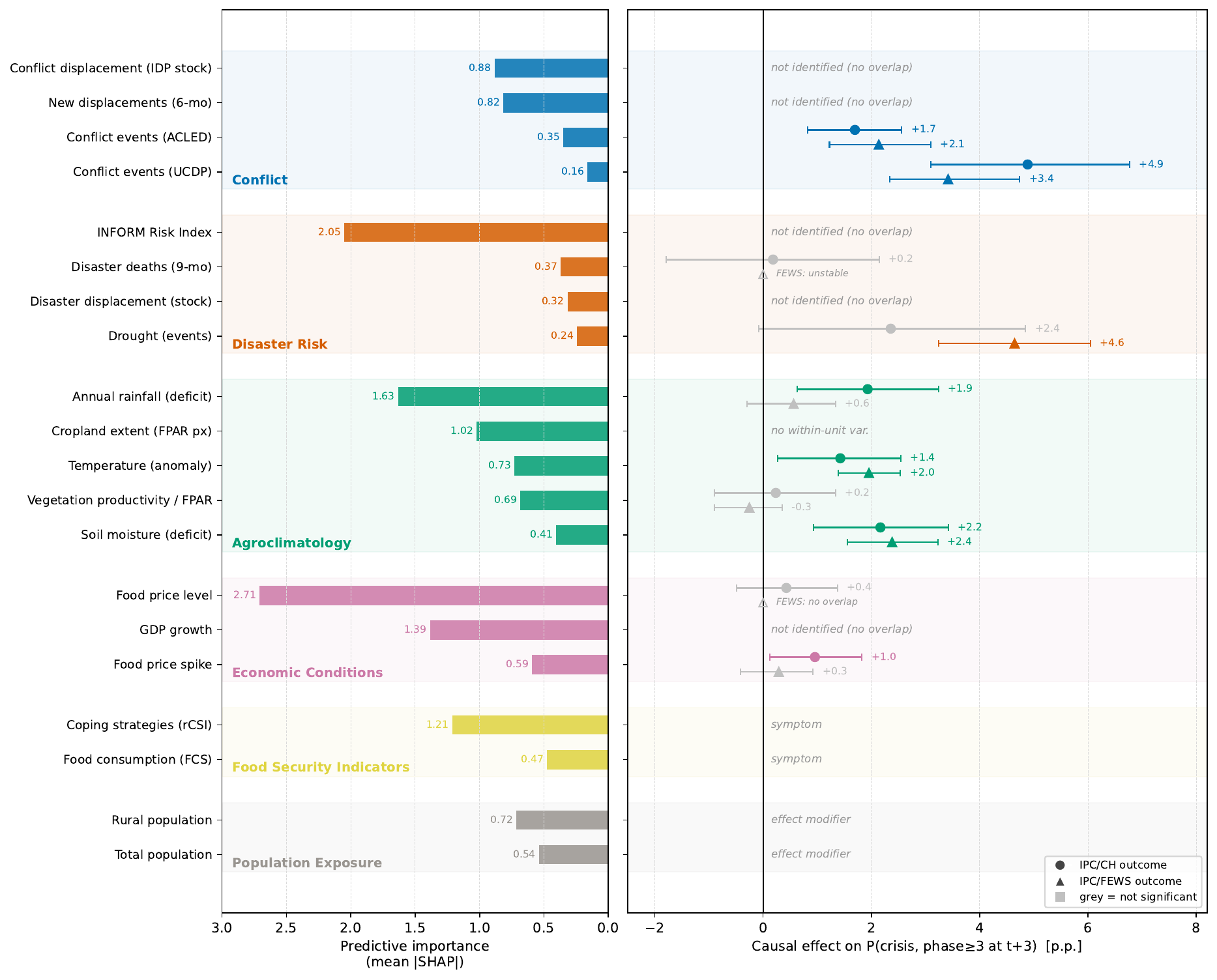}
\caption{Predictive importance does not track causal effect. Per model feature (grouped by driver thematic dimension): predictive importance (mean |SHAP|, left) vs the Average Treatment Effect on crisis probability (IPC/CH $\geq$ 3, t+3) of a within-unit $\pm$1 SD adverse shock. Coloured = significant, grey = non-significant; features with no common support (displacement stocks, INFORM, GDP) are ``not identified,'' and food-security indicators and population are downstream symptoms/exposure, not treatments.}
\label{fig:causal}
\end{figure}

\subsection*{Predictive signals diverge from causal effects}


Overall, we observe that predictive feature attribution and estimated causal effects ($\tau_{\text{AIPW}}$) diverge sharply, a pattern that remains robust across both classification systems (see Fig. \ref{fig:causal}, Supplementary Table \ref{tab:ipc_drivers} and Supplementary Table \ref{tab:fews_drivers}). The top global forecasting feature, the local food-price level (mean |SHAP| = 2.71), yields no detectable causal effect on crisis probability three months later. The second, the INFORM Risk Index, is non-identifiable: as a slow-moving structural baseline, it varies too little within-unit to be distinguished from the fixed effects. This predictive-causal decoupling is equally pronounced within individual thematic dimensions. Within the Conflict domain, the main forecasting anchor (cumulative stock of internally displaced persons) is likewise non-identifiable: measured only at country-level, it has no within-country variation beyond the fixed effects. The active causal signal lies instead in acute conflict shocks, which ranked lowest in unified predictive importance (ACLED events: $+1.7$ p.p., $95\%$ CI $0.8\text{--}2.6$; UCDP events: $+4.9$ p.p., $95\%$ CI $3.1\text{--}6.8$; both robust under FEWS NET). Within Economic Conditions, the causal impact comes entirely from localized food price spikes ($+1.0$ p.p., $95\%$ CI $0.1\text{--}1.8$), rather than absolute baseline price levels or lagged macro-structural GDP growth. 

Agroclimatology is the only dimension where the main predictive feature aligns with an identifiable causal lever: acute rainfall deficit ($+1.9$ p.p., $95\%$ CI $0.6\text{--}3.2$), soil moisture deficits ($+2.2$ p.p.), and temperature anomalies ($+1.4$ p.p.) all significantly elevate crisis probability, whereas the model's secondary vegetation-productivity proxy (FPAR) yields no independent marginal effect. Crucially, these effects arise only from severe shocks ($\pm 1$ SD binarized shocks), as the continuous per-SD linear gradients are not significant. Disaster Risk variables are largely non-identifiable, except drought events: despite weak predictive importance, they show a significant causal effect under FEWS NET (+4.6 p.p., 95\% CI 3.2–6.0), echoing that low-SHAP features can carry substantial causal weight. The remaining indicators fall outside an intervention framework: food-security indicators (rCSI, FCS) represent downstream symptoms of distress, while population variables act as exposure scales or effect modifiers, not levers. These identified causal shocks are modest and only moderately robust to unmeasured confounding (E-values of $\approx 1.2\text{--}1.4$ under IPC/CH and $1.5\text{--}2.0$ under FEWS NET). 

Overall, unified SHAP importance reflects cross-sectional predictive utility blending causal effects with baseline confounding and downstream proxy signals. Consequently, SHAP rankings should not be read as a hierarchy of causal levers; acute food crises are driven less by the structural baselines the model uses to sort global risk than by sudden shocks due to conflict, local market prices, and climate.

\section*{Discussion}

Food crisis classification forecasting has long been constrained by limited geographic scope, single-system validation, and underdeveloped evaluation protocols. This study addresses these gaps at scale, and further examines what predictive attributions from such models can—and cannot—reveal about the underlying drivers, clarifying the boundary between predictive utility and causal understanding.


Temporally nested cross-validation yields more robust generalization estimates than the single held-out periods common in prior works \cite {machefer2025potential, van2024unsupervised}, capturing performance variance over time and enabling principled uncertainty quantification. The expanded HFID dataset broadens geographic coverage well beyond existing approaches, which reach at most 21 countries and rely almost exclusively on FEWS NET\cite{andree2020predicting}. Missing-data-resilient models (HistXGBoost, TabPFN) capture sufficient predictive signal from driver features alone: adding the lagged target improves performance by at most 7 percentage points across all metrics on the ``Full dataset'', highlighting direct benefits in where  prior assessments are unavailable \cite{wang2022transitions, westerveld2021forecasting, busker2024predicting}.

Cross-country performance differences are driven mainly by positive class prevalence, not data volume or classification system. Cross-source discrepancies, such as Afghanistan's PR-AUC of 0.963 under IPC/CH versus 0.404 under FEWS NET, likewise reflect prevalence differences arising from each system’s temporal coverage and phase attribution, not model failure. A single global model is therefore unlikely to capture crisis dynamics across all prevalence profiles. Profile-specific models and rare-event methods, such as cost-sensitive learning or synthetic data augmentation \cite{shyalika2024comprehensive}, are the most promising ways to improve low-prevalence performance and could be combined in a unified system through ensembling.

The SHAP-based analysis confirms that food crisis classification prediction relies on compound, geography-specific signals rather than universal leading indicators. Economic conditions and agroclimatology serve as near-universal baseline dimensions across all forecasting archetypes, whereas conflict and disaster risk are concentrated in specific regional contexts. Three recurrent archetypes emerge — agroclimatic-structural, market-dependent dominant, and conflict-agricultural — each stable over time and consistent with known regional crisis dynamics \cite{fsin2024grfc}. The model's capacity to capture acute temporal restructuring, including a 79\% month-on-month acceleration in conflict attribution in Palestine following October 2023 and a sharp acceleration in economic attribution in Ukraine from January 2024, suggests that SHAP-based attribution profiles could support near-real-time monitoring of shifts in crisis trajectories.

The most policy-relevant finding is that predictive importance and causal effect systematically diverge. This divergence is largely expected by design: SHAP rewards stable predictors, whereas causal estimation requires variation. The strongest risk-sorting features — baseline price levels, displacement stocks, structural risk indices — are largely non-actionable within three-month, whereas conflict events, food prices spikes, and agroclimatic shocks, are both causally significant and actionable. Treating SHAP rankings as intervention prorities risks misallocating resources toward structural baselines that improve forecasting but misinform short-term intervention policy. The two analyses are complementary: SHAP for forecast interpretation and regional profiling, causal estimates for targeting anticipatory action.

Several limitations warrant attention. Emergency occurrence and crisis escalation remain difficult to model reliably due to severe class imbalance. Causal estimates are structurally modest and only moderately robust to unmeasured confounding, with E-values of 1.2--1.4 under IPC/CH and 1.5--2.0 under FEWS NET. Closer collaboration with humanitarian practitioners will also be essential to identify metrics that best reflect operational decision-making needs and to ensure that model outputs are well calibrated and actionable for prevention.

More broadly, this framework could underpin an operational decision-support tool for humanitarian organizations with forecasts triggering early financing mechanisms, and causal estimates prioritizing conflict, market, and climate response pathways. This requires co-design with operational partners (WFP, FAO, and FEWS NET) along with transparency, auditability, and integration into existing early warning workflows. As humanitarian needs grow and budgets shrink, scalable and causally grounded forecasting offers a concrete step toward more efficient and targeted anticipatory action. 


\section*{Methods}

\subsection*{Expanded HFID dataset} 

\subsubsection*{Food Security Indicators}
The Harmonized Food Insecurity Dataset (HFID)\cite{machefer2025monthly} spans Global Administrative Areas (GADM) levels 1 and 2 (Admin-1/Admin-2) across 80 countries and provides monthly data, when and where available, from June 2007 to May 2024, drawing from four primary sources of food insecurity indicators: IPC/CH phases, FEWS NET IPC-compatible phases, WFP's prevalence of insufficient Food Consumption Score (FCS) and Reduced Coping Strategies Index (rCSI). In this study, IPC/CH phases and FEWS NET IPC-compatible phases are used as target variables, while the statistics (monthly minimum, maximum, mean) on prevalence of insufficient FCS and rCSI is used as input features capturing downstream manifestations of food insecurity.  As noted in Results, these indicators are not treated as intervention targets in the causal analysis, as they reflect behavioral and consumption responses to active distress rather than upstream drivers.

None of these indicators shares the same spatio-temporal coverage. The HFID is further expanded here to include driver variables for all possible year-month combinations for each administrative area, not only when food insecurity indicators are available. Details on dataset coverage are given in Supplementary Table \ref{tab:hfid_variables}.

\subsubsection*{Economic Conditions}
We use food price data from two complementary sources: the World Bank's Food Prices dataset\cite{andree2021monthly}, which provides monthly market-level data for 36 countries (accessed 12/2024), and the FAO's global daily food price indices\cite{faoffpi}. The World Bank dataset offers machine-learning-based estimates (open, high, low, and close) of a composite food price index from 2007 onward, capturing subnational price dynamics — particularly in rural or vulnerable areas often overlooked by traditional aggregate statistics. Market prices are averaged at the Admin-2 level, with national averages serving as an approximation for unmatched subnational areas. Remaining missing values are imputed using a temporal gap-filling strategy: linear interpolation for internal gaps, forward fill for values missing at the end of the time series, and backward fill for values missing at the beginning. A three-month rolling moving average and z-score normalization are subsequently applied. The FAO Food Price Index, available from 1990 onward, is monthly averaged and gap-filled by linear interpolation, along with its component indices for Meat, Dairy, Cereals, Oils, and Sugar.

Annual GDP growth data are obtained from the World Bank\cite{wbgdp}, covering 184 countries over 1961--2023 (accessed 12/2025). The data are provided at the country-year level and measure the percentage change in real GDP relative to the previous year. A two-year lag is applied: for any given observation year, the GDP growth rate recorded two years prior is assigned. This serves two purposes: GDP growth data for a given year are often unavailable until one or two years later, and using lagged economic conditions reduces the risk of reverse causality. Missing values are filled using the temporal gap-filling strategy described above. 

Because gap-filling and normalization were applied to the full dataset rather than within each training fold, imputed values may incorporate information observed after the prediction date; given the short gaps and smooth evolution of these economic series, we expect this leakage to be minor, but we acknowledge it as a limitation.

\subsubsection*{Agroclimatology}
Meteorological drivers and Earth Observation (EO) data at 10-day temporal resolution were entirely sourced from the JRC Anomaly HotSpots of Agricultural Production (ASAP \footnote{https://agricultural-production-hotspots.ec.europa.eu/}) early warning system\cite{rembold2019, fritz2019, rembold2023, meroni2025}. ASAP global meteorological variables include precipitation and air temperature, sourced from the Climate Hazards Group InfraRed Precipitation with Station data (CHIRPS) product for latitudes between 50°S and 50°N at 0.05° spatial resolution; outside this band, precipitation is obtained from the European Centre for Medium-Range Weather Forecasts (ECMWF) ERA5 and ENS models at 0.25° spatial resolution. Air temperature at 0.25° spatial resolution is also retrieved from ECMWF. Near-real-time estimates are generated by combining ERA5 reanalysis and ECMWF deterministic ENS forecast data\cite{meroni2025}. ASAP global EO variables include the fraction of absorbed photosynthetically active radiation (FPAR) and surface soil moisture. FPAR at a 500 m spatial resolution is derived by blending MODIS observations (2000--2018) and VIIRS observations thereafter \cite{seguini2025}. Volumetric surface soil moisture at 25 km spatial resolution, derived from combined active and passive microwave satellite observations, is obtained from the Copernicus Climate Data Store.

To reduce the dimensionality of the resulting 10-day time series, features are derived as weighted temporal averages tailored for AFI estimation, representing recent FPAR and agroclimatic anomalies that may influence local food production. Because AFI forecasting may be triggered at any time of the year, the temporal window considered must span a full solar year or a multiple thereof; shorter windows (e.g., three or six months) could capture periods of limited agronomic relevance, such as dry seasons when crops are absent. Accordingly, two temporal windows are considered: one year and two years, with the latter applied only to FPAR, extending further into the past to capture potential crop or rangeland failures from previous growing seasons that may still influence current household food insecurity. Within each window, greater weight is given to periods when crops or rangelands are climatologically active and to more recent observations. This is achieved by multiplying two weighting factors: (i) the climatological fraction of active crop or rangeland area for each 10-day period of the year, derived from ASAP and based on satellite-derived multi-annual mean phenology, and (ii) a linear backward-decreasing function of time, equal to 1 at the forecast date and decreasing to 1/36 and 1/72 at the earliest 10-day step of the one-year and two-year windows, respectively (36 being the number of 10-day periods per year). For precipitation and soil moisture, the active-fraction time series is shifted forward by one month to reflect the agronomic relevance of pre-season moisture conditions and the reduced sensitivity of crops to these variables toward the end of the growing season. The final weighted-average feature for each variable is computed as the weighted temporal average over the selected window, using the product of the two weighting factors, and is standardized at the administrative-unit level.

\subsubsection*{Population Exposure}
Subnational population data are sourced from the Global Human Settlement Layer (GHSL), using the 2005, 2010, 2015, 2020, and 2025 releases of the GHS-DUC product, which provides urbanization classifications for administrative units following the Stage II Degree of Urbanization methodology. Total population and its breakdown by urban centers, clusters, and rural categories are extracted. Each HFID record is matched to the population release closest in time at the Admin-1 or Admin-2 level, where available.

\subsubsection*{Conflict}
Conflict data is drawn from two main sources: the Uppsala Conflict Data Program (UCDP) and the Armed Conflict Location \& Event Data Project (ACLED), which provide event-level records of fatalities, with start and end dates, and locations. Events and fatalities are aggregated to the monthly level — with events spanning multiple months counted in each relevant month — and spatially aggregated to subnational boundaries. Conflict drivers are merged across the full temporal coverage of the HFID. The missing values are filled with zero for UCDP, which provides global coverage from 1989 onward, and similarly for ACLED \cite{acledccov}, but only from the date each country started to be covered.

Displacement data are sourced from the Internal Displacement Monitoring Centre (IDMC) Global Internal Displacement Database (GIDD), tracking internal displacement due to conflict and generalized violence (since 2009) and sudden-onset natural disasters (since 2008). The GIDD provides national-level information on population flows (new displacements occurring during the year) and stocks (total people living in displacement at year-end). IDMC data are aligned with the HFID at the country and year levels; missing entries are filled with zero. Monthly displacement data at the national level\cite{rost2025anticipating} (from 2019 onward) are also included, smoothed using rolling sums over 3-, 6-, and 9-month windows.

\subsubsection*{Disaster Risk}
Flood, storm, drought, and mixed disaster events\footnote{Road, Flood, Extreme temperature, Fire (Miscellaneous), Volcanic activity, Storm, Wildfire, Earthquake, Rail, Air, Collapse (Industrial), Collapse (Miscellaneous), Fire (Industrial), Explosion (Miscellaneous), Epidemic, Water, Mass movement (wet), Explosion (Industrial), Chemical spill, Gas leak, Infestation, Miscellaneous accident (General), Poisoning, Mass movement (dry), Industrial accident (General), Radiation, Oil spill, Impact, Animal incident, Glacial lake outburst flood} are obtained from a recent work\cite{ronco2025subnational} that derived monthly subnational (Admin-1) data from EM-DAT, the International Disaster Database. Monthly rolling sums over the last three, six, and nine months are also computed. A vulnerability index for each month of the year is computed by summing the number of events occurring before 2020 for each event type. We also integrate the INFORM Risk Index, a global open-source assessment that annually measures the risk of humanitarian crises and disasters at the country level. Each HFID record is matched with the previous year's INFORM Risk Index to ensure only historical information is used.




\subsection*{Models training} 
\subsubsection*{Settings}

We consider state-of-the-art models explored in previous studies \cite{machefer2025potential}, such as Logistic Regression, Decision Tree, Random Forest, and XGBoost. We also include a Histogram-Based Gradient Boosting (HistXGBoost) and TabPFN \cite{hollmann2025} (v2 default version), both more recent approaches that are faster and more memory-efficient methods and have native support for missing values.  

We evaluate models across the following configurations:
\begin{itemize}
    \item Target source: either FEWS NET or IPC/CH;
    \item Target type: all binary targets - crisis occurrence (IPC Phase >3), emergency occurrence (IPC Phase >4), and
crisis escalation (when a region either transitions from no crisis into crisis or deepens an already ongoing crisis).
    \item Lead time: 3, 6, 9 months;
    \item Previous target available: whether or not the last available target value is included as a feature. When included, we test regular forecasting; when excluded, we estimate the capacity of the model when previous assessments are not available; 
    \item Model: Logistic Regression, Decision Tree, Random Forest, XGBoost, HistXGBoost, TabPFN. Each pipeline is trained with different hyperparameters (see Supplementary Table \ref{tab:hyperparam_grid}).
    \item Dataset completeness: we consider two training regimes, a ``Restricted'' dataset that retains only fully complete observations, and a ``Full'' dataset that includes nearly all observations by admitting those with up to 20\% missing features per observation;
    \item Evaluation subset: metrics are computed either over all administrative areas or restricted to non-constant ones (NC) that experience temporal target variation.
\end{itemize}

\subsubsection*{Temporal nested cross-validation strategy}

We employ a temporal nested cross-validation (Supplementary Figure \ref{fig:cval_timeline}). In the outer loop, a fixed 4-month test window advances sequentially through time, so that each observation appears in exactly one outer test fold; aggregating predictions across folds yields an unbiased estimate of generalization performance and allows uncertainty quantification from the distribution of fold-level metrics.  The 4-month window is chosen to ensure that every test fold contains observations: FEWS NET releases assessments at least every 4 months, while IPC/CH analyses are available monthly, albeit from different countries \cite{machefer2025potential}. An inner loop of 3 folds — each with a 4-month validation window, likewise respecting temporal ordering — is used exclusively for hyperparameter optimization and feature selection, so that outer test performance is never influenced by model-selection decisions. Because the training set expands across folds, the scheme mimics operational deployment, where growing archives of past observations are the only information available at prediction time (see Supplementary Table \ref{tab:cv_summary} for summary statistics of the cross-validation datasets). Feature selection is done by ranking features with the ``Minimum Redundancy - Maximum Relevance'' algorithm \cite{ding2005minimum}, and the number of retained features is treated as a hyperparameter in the inner loop (Supplementary Table \ref{tab:hyperparam_grid}).

\subsubsection*{Performance metrics and statistical significance}

Given the binary classification setting and the class imbalance present in this study, we evaluate model performance using several complementary metrics. Accuracy measures the overall proportion of correct predictions, while the F1-score captures the harmonic mean of precision and recall, providing a more balanced assessment when classes are unevenly distributed. Where the F1-score is computed at the optimal threshold, a country-level cutoff is derived from the test data, rendering it a retrospective measure of classification performance. 

The Precision-Recall Area Under the Curve (PR-AUC) summarizes model discrimination across all binarization thresholds and is particularly well-suited to imbalanced settings, as it focuses on the positive (minority) class without being inflated by true negatives. To assess probabilistic calibration — that is, how well predicted probabilities reflect true event frequencies — we also report the Brier Score, which measures the mean squared error between predicted probabilities and observed binary outcomes; lower values indicate better-calibrated, more accurate probabilistic predictions.

To assess whether the observed performance differences between pipelines are statistically meaningful \cite{rainio2024evaluation}, we apply nonparametric significance tests at the 0.05 significance level. The Wilcoxon signed-rank test is used for pairwise comparisons between two models, testing whether their performance distributions differ systematically across folds. When comparing multiple pipelines simultaneously, we apply the Friedman test — a rank-based alternative to repeated-measures ANOVA — followed by the Nemenyi post-hoc test \cite{nemenyi1963distribution} to identify which pairs of models differ significantly while controlling for multiple comparisons. We provide the average rank of each model to explicitly indicate the best one. 

\subsubsection*{Naive baseline and evaluation scope}

All top-performing models are evaluated against a naive persistence baseline that outputs the most recent available target value. To isolate where ML models add genuine predictive value, this comparison is restricted to administrative areas experiencing temporal target variation (described as Non-Constant ``NC''), where the naive persistence baseline cannot, by definition, capture genuine transitions. Because the naive baseline yields deterministic binary predictions rather than calibrated probabilities, its Brier Score reduces to the raw misclassification rate, giving 
probabilistic models an inherent calibration advantage. Since it produces no probabilistic ranking, a PR-AUC cannot be computed directly; it is therefore benchmarked against a no-skill baseline defined by the positive-class prevalence \cite{saito2015precision}.

\subsection*{Models explainability analysis}

\begin{table}[h]
\centering
\begin{tabular}{|l|p{10cm}|}
\hline
\textbf{Thematic dimension} & \textbf{Variables} \\ \hline
Conflict & Conflict displacement (IDP stock), New displacements (6-mo), Conflict events (ACLED), Conflict events (UCDP) \\ \hline
Disaster Risk & INFORM Risk Index, Drought (events), Disaster displacement (stock), Disaster deaths (9-mo) \\ \hline
Agroclimatology & Vegetation productivity / FPAR, Annual rainfall (deficit), Soil moisture (deficit), Temperature (anomaly), Cropland extent (FPAR px) \\ \hline
Economic Conditions & GDP growth, Food price level, Food price spike \\ \hline
Food Security Indicators & Coping strategies (rCSI), Food consumption (FCS) \\ \hline
Population Exposure & Rural population, Total population \\ \hline
\end{tabular}
\caption{Input features grouped into six thematic dimensions used for predictive and causal attribution analysis.}
\label{tab:feature_categories}
\end{table}

To map the compound predictive signals of acute food crises independently of any single classification system, we synthesized the feature attributions of two HistXGBoost classifiers (i.e., crisis occurrence for FEWS NET and IPC/CH, respectively). Local feature attributions were computed using TreeSHAP \cite{lundberg2017unified} via an interventional feature perturbation approach backgrounded on the training data, with the model output explicitly configured in probability space:$$\phi_{i,f} \in \mathbb{R}$$where $\phi_{i,f}$ represents the change in predicted crisis probability attributed to feature $f$ for observation $i$. While deep tabular foundation models such as TabPFN support post-hoc interpretability via model-agnostic adaptations (e.g., KernelSHAP), calculating feature attributions for a transformer-based model requires computationally intensive in-context evaluations across exponential feature coalitions. In contrast, HistXGBoost interfaces natively with TreeSHAP, a highly optimized algorithm that computes exact Shapley values, reducing the computational complexity from exponential (for general models) to polynomial time by directly exploiting internal tree structures\cite{lundberg2020local}. This structural efficiency ensures exact attribution tracking without the sampling variance or prohibitive computational overhead that model-agnostic permutation frameworks introduce. This calculation was executed independently for the FEWS NET and IPC/CH HistXGBoost models at a three-month predictive lead time. Because the two classifiers operate on distinct underlying optimization surfaces, their raw SHAP values exhibit differences in scale. To ensure mathematical comparability while preserving the relative importance hierarchy of features within each model, we normalized the raw SHAP matrices by their global mean absolute magnitude across all observations (N) and features (F):$$S_{\text{FEWS}} = \frac{1}{N \cdot F} \sum_{j=1}^{N} \sum_{f=1}^{F} |\phi_{j,f,\text{FEWS}}|$$$$S_{\text{IPC/CH}} = \frac{1}{N \cdot F} \sum_{j=1}^{N} \sum_{f=1}^{F} |\phi_{j,f,\text{IPC/CH}}|$$The scaled SHAP values were then defined as:$$\hat{\phi}_{i,f,\text{FEWS}} = \frac{\phi_{i,f,\text{FEWS}}}{S_{\text{FEWS}}}, \quad \hat{\phi}_{i,f,\text{IPC/CH}} = \frac{\phi_{i,f,\text{IPC/CH}}}{S_{\text{IPC/CH}}}$$To account for differences in model discrimination performance, the scaled attributions were combined via a weighted average proportional to each classifier's Average Precision ($AP$) score evaluated on the holdout test set. The normalized weights $w_{\text{FEWS}}$ and $w_{\text{IPC/CH}}$ were derived as:$$w_{\text{FEWS}} = \frac{AP_{\text{FEWS}}}{AP_{\text{FEWS}} + AP_{\text{IPC/CH}}}, \quad w_{\text{IPC/CH}} = \frac{AP_{\text{IPC/CH}}}{AP_{\text{FEWS}} + AP_{\text{IPC/CH}}}$$Yielding the final unified attribution signal (shap importance) $\Phi_{i,f}$ for each observation:$$\Phi_{i,f} = w_{\text{FEWS}} \cdot \hat{\phi}_{i,f,\text{FEWS}} + w_{\text{IPC/CH}} \cdot \hat{\phi}_{i,f,\text{IPC/CH}}$$To isolate the predictive profiles specific to acute humanitarian stress, the subsequent interpretability analysis was strictly restricted to a union-based panel of predicted crisis states. A spatiotemporal observation ($\text{Admin-2} \times \text{month}$) was included in the evaluation set if either the FEWS NET or the IPC/CH classifier predicted a crisis phase or worse ($\geq 3$). 


To facilitate a higher-level structural analysis of the forecasting model's behavior, individual feature attributions were aggregated into six established thematic dimensions: Agroclimatology, Conflict, Economic Conditions, Food Security Indicators, Disaster Risk, and Population Exposure, following Machefer et al. \cite{machefer2025potential} (see Table \ref{tab:feature_categories} for full variable mapping). To preserve the additive efficiency inherent to Shapley values, global thematic dimension scores are derived by summing the mean absolute unified SHAP importance of all constituent features within each group. Let $\mathcal{M}: f \to C$ map each individual feature $f$ to its respective thematic dimension $C$. The global attribution score $\Omega_C$ for category $C$ across the evaluation set of $N$ crisis observations is computed as:$$\Omega_C = \sum_{f \in \mathcal{F}_C} \left( \frac{1}{N} \sum_{i=1}^{N} |\Phi_{i,f}| \right)$$where $\mathcal{F}_C = \{f \mid \mathcal{M}(f) = C\}$ represents the complete subset of features assigned to category $C$. Because individual TreeSHAP values reflect local additive contributions in probability space, summing their mean absolute magnitudes guarantees that the system's total predictive weight of the system is efficiently conserved and fully accounted for across the six macro-dimensions. 


To map recurrent subnational forecasting archetypes, we analyze the static attribution profile of each administrative unit. Unified feature attributions are first temporally averaged over the evaluation window for each unique $\text{Admin-2}$ unit. To decouple the structural composition of these signals from their absolute scale, we apply row-wise min-max normalization that maps the lowest-impact category to $0$ and the highest to $1$ for each row. Direct $K$-means clustering ($K=3$, $n_{\text{init}}=100$) is then applied to these normalized internal shapes, classifying the administrative units based on their underlying semantic configuration rather than on absolute crisis severity. The resulting cluster centroids are reported as both absolute mean SHAP profiles and relative percentage recipes.

To populate the global geospatial visualization (Figure~\ref{fig:shap_map}), the dominant thematic dimension and its accompanying growth-intensity shading are determined at the Admin-2 level by combining a magnitude term with a directional trend term. For each chronologically sorted administrative unit time series, the evaluation window is segmented into a baseline phase and a recent phase using a window size of \(m = 6\) months, following standard baseline-versus-recent comparison approaches used in trend and change detection. A local trend component \((\Delta_{u,C})\) is computed as the difference between the recent-phase and baseline-phase means, with a rectified (ReLU-style) floor at zero so that only growing or stable signals contribute, filtering out decaying trajectories. This is combined multiplicatively with a magnitude component \((\bar{\Omega}_{u,C})\), defined as the overall mean absolute presence of category \(C\) in unit \(u\) across the full window, yielding a composite score:

\[
\mathrm{Score}_{u,C} = \bar{\Omega}_{u,C} \cdot \Delta_{u,C}
\]

This construction follows the common approach of weighting a signal's magnitude by its recent growth to prioritize categories that are both substantial in absolute terms and actively intensifying, rather than those that are merely large but static or declining. For each unit, the mapped category corresponds to the domain \(C\) maximizing \(\mathrm{Score}_{u,C}\), and this score determines the color saturation on the map (Figure~\ref{fig:shap_map}). If the score is zero or negative across all categories for a given unit — indicating flat or uniformly decaying trajectories — the engine defaults to the category with the highest absolute presence, \(\bar{\Omega}_{u,C}\).

Finally, to evaluate structural shifts in model reliance over time, we construct a global temporal deviation matrix tracking monthly changes in category importances. For each category and month, we calculate the percentage deviation of the cross-sectional mean absolute SHAP value relative to its global temporal baseline over the entire study horizon. The resulting matrix is visualized via a temporal heatmap. To ensure visual stability against outliers while preserving a symmetrical, zero-anchored color bar, the color limits are determined symmetrically by taking the maximum of the absolute values of the 5th and 95th percentiles of the global deviation distribution.

\subsection*{Causal inference analysis}

To isolate actionable policy levers from mere predictive markers, we formalize our causal framework using the potential outcomes notation. Let $D_{i,t} \in \{0, 1\}$ denote a binary adverse ``shock'' to a given feature in subnational unit $i$ at month $t$, and let $Y_{i,t+3} \in \{0, 1\}$ represent the binary outcome indicating whether the unit enters a crisis or worse phase ($\geq 3$) three months ahead. Each observation has two potential outcomes, $Y_{i,t+3}(1)$ and $Y_{i,t+3}(0)$, corresponding to the crisis state under treatment (shock) and control (typical conditions), respectively. Our primary estimand is the Average Treatment Effect (ATE), defined as\cite{Chernozhukov2016}:$$\text{ATE} = \mathbb{E}[Y_{i,t+3}(1) - Y_{i,t+3}(0)]$$We estimate the ATE using cross-fitted augmented inverse-propensity weighting (AIPW), a doubly robust estimator that consistently evaluates the causal effect if either the propensity score model, $e(X) = P(D=1 \mid X)$, or the outcome regression model, $\mu_d(X) = \mathbb{E}[Y \mid D=d, X]$, is correctly specified. The AIPW estimand for the sample is formulated as:$$\tau_{\text{AIPW}} = \frac{1}{N}\sum_{i=1}^{N} \left[ \left( \mu_1(X_i) - \mu_0(X_i) \right) + \frac{D_i \left(Y_i - \mu_1(X_i)\right)}{e(X_i)} - \frac{(1 - D_i)\left(Y_i - \mu_0(X_i)\right)}{1 - e(X_i)} \right]$$where $X_i$ represents the baseline covariates within the variable-specific back-door adjustment set. We estimated this average treatment effect for each model feature on crisis-or-worse occurrence, defined as IPC/CH (and, as a robustness check, FEWS NET) phase $\geq$ 3 three months ahead, at the Admin-2 monthly level on the expanded HFID. Each continuous feature was 
converted to a binary adverse ``shock'' by standardizing it within administrative units and thresholding one-sided at 1 SD in the adverse direction (z $\leq -1$ for protective variables such as rainfall, z $\geq +1$ for harmful variables such as conflict or prices), contrasting an unusually adverse month with all other months for the same location. Effects were estimated by cross-fitted AIPW with gradient-boosted outcome and propensity models, country/year/month fixed effects, and propensity clipped to [0.05, 0.95]; confidence intervals are from a nonparametric bootstrap clustered at Admin-1. Each feature used its own back-door adjustment set, derived from an explicit causal graph, conditioning on common causes while excluding mediators and colliders: weather shocks (rainfall, soil moisture, temperature, drought), being effectively exogenous, were adjusted only for baseline severity, population and fixed effects (not conflict, which lies on the weather$\rightarrow$conflict$\rightarrow$crisis path), whereas conflict events were adjusted for weather, and prices for conflict, weather and GDP. For every estimate, we report a placebo (permuted-treatment) test, a common-support diagnostic (share of observations with propensity outside [0.05, 0.95]), a continuous per $\pm$1 SD dose-response (partial-linear double ML), and an E-value for unobserved-confounding sensitivity. Estimates without common support ($\geq$ 95$\%$ off-support: cumulative stocks such as conflict/disaster displacement, and slow-moving aggregates such as INFORM and GDP, whose within-unit variation is collinear with the fixed effects) are reported as non-identifiable rather than as effects. We additionally apply a stability screen: a significant estimate whose sign reverses relative to the unadjusted association and whose placebo exceeds 0.3 p.p. is treated as non-identifiable (this flags only FEWS-NET disaster deaths). Features that do not admit a well-defined intervention were excluded: food-security indicators (rCSI, FCS) are downstream measures of food insecurity, and population variables act as exposure and effect modifiers.

\subsection*{Data Availability} 

All data needed to support the conclusions in the paper can be downloaded from \href{https://doi.org/10.5281/zenodo.24281248}{\url{https://doi.org/10.5281/zenodo.24281248}}. 

\subsection*{Code Availability} 

The code to reproduce all experiments and results can be downloaded from \href{https://doi.org/10.5281/zenodo.24140289}{\url{https://doi.org/10.5281/zenodo.24140289}}.

\bibliography{sample2}
\bibliographystyle{naturemag-doi}

\section*{Acknowledgements}

MR and MMa would like to thank Duccio Piovani (WFP) for providing curated data on food consumption and coping strategies, and for his continuous support, meetings, and careful review of the manuscript. Special thanks also go to Giulia Martini (WFP) for fruitful discussions on food security data and modeling, and for reading an early version of this work. MR and MMa thank Andree Bo (World Bank) for his early-stage feedback, support, and discussions on food prices. MR thanks Jerome Stephen White (World Bank) for discussions on food crisis anticipation and the use of non-standard data sources as signals. MR is grateful to Ioannis Prapas for fruitful conversations on the topic.

\section*{Author contributions statement}

MMa: Conceptualization, Data curation, Formal analysis, Software, Writing – original draft preparation; MR: Conceptualization, Data curation, Formal analysis, Software, Writing – original draft preparation; MMe: Methodology, Writing – original draft preparation; JCB: Methodology, Data curation, Formal analysis, Software, Writing – original draft preparation; FC: Data curation; CK: Writing – review \& editing; NR: Writing – review \& editing; AM: Writing – review \& editing; ERB: Writing – review \& editing;   PZ: Writing – review \& editing; GCV: Supervision, Methodology, Writing – original draft preparation; CC: Supervision; FR: Supervision;

\section*{Competing Interests Statement}

The authors declare no competing interests.

\section*{Disclaimer}
The views expressed in this article are those of the authors and do not necessarily reflect the views of the United Nations.
\clearpage
\section*{Supplementary information}

\setcounter{figure}{0}
\renewcommand{\thefigure}{S\arabic{figure}}

\setcounter{table}{0}
\renewcommand{\thetable}{S\arabic{table}}

\subsubsection*{Expanded HFID dataset records}

\label{exp_hfid_cov}
The expanded HFID contains 83 driver variables and 2 target variables. Each entry is further annotated with country identifiers (name, ISO-2, ISO-3), the corresponding UN Geoscheme region, and a year-month identifier. Not all food insecurity indicators in the HFID share the same spatio-temporal coverage, whether for the driver variables (prevalence of insufficient FCS and rCSI), the target variables (IPC/CH and FEWS NET IPC-compatible phases), or the additional drivers used to expand the dataset across all year-month combinations for each administrative area. The coverage of the dataset is further detailed in Supplementary Table \ref{tab:hfid_variables}.

\begin{longtable}{p{7cm}p{1.7cm}p{1.7cm}p{1.7cm}p{1.1cm}p{1.1cm}p{1.1cm}}
\caption{Spatio-temporal coverage statistics and missing data rate for the variables in the expanded HFID. Number (N) of administrative area (adm) covered are provided at level 0,1 and 2.} \label{tab:hfid_variables} \\
\toprule
Variable & From & To & Missing (\%) & N adm0 & N adm1 & N adm2 \\
\midrule
\endfirsthead
\caption[]{Spatio-temporal coverage statistics and missing data rate for the variables in the expanded HFID. Number (N) of administrative area (adm) covered are provided at level 0,1 and 2.} \\
\toprule
Variable & From & To & Missing (\%) & N adm0 & N adm1 & N adm2 \\
\midrule
\endhead
\midrule
\multicolumn{7}{r}{Continued on next page} \\
\midrule
\endfoot
\bottomrule
\endlastfoot
ipc\_phase\_fews & Jan 2011 & Feb 2024 & 81\% & 28 & 483 & 3911 \\
ipc\_phase\_ipcch & Jan 2017 & May 2024 & 87\% & 49 & 678 & 5022 \\
fcs\_rt mean & Oct 2018 & Apr 2024 & 83\% & 33 & 494 & 3302 \\
fcs\_rt max & Oct 2018 & Apr 2024 & 83\% & 33 & 494 & 3302 \\
fcs\_rt min & Oct 2018 & Apr 2024 & 83\% & 33 & 494 & 3302 \\
rcsi\_rt mean & Oct 2018 & Apr 2024 & 83\% & 33 & 494 & 3302 \\
rcsi\_rt max & Oct 2018 & Apr 2024 & 83\% & 33 & 494 & 3302 \\
rcsi\_rt min & Oct 2018 & Apr 2024 & 83\% & 33 & 494 & 3302 \\
px\_cnt\_weight\_sum\_FPAR\_crop & Jun 2007 & May 2024 & 12\% & 80 & 1209 & 4810 \\
px\_cnt\_weight\_sum\_FPAR\_past & Jun 2007 & May 2024 & 2\% & 80 & 1242 & 5429 \\
w\_avg\_1y\_FPAR\_crop & Jun 2007 & May 2024 & 25\% & 78 & 1139 & 4046 \\
w\_avg\_1y\_FPAR\_past & Jun 2007 & May 2024 & 5\% & 80 & 1228 & 5255 \\
w\_avg\_1y\_RAIN\_crop & Jun 2007 & May 2024 & 25\% & 78 & 1139 & 4046 \\
w\_avg\_1y\_RAIN\_past & Jun 2007 & May 2024 & 5\% & 80 & 1228 & 5255 \\
w\_avg\_1y\_SM\_combined\_crop & Jun 2007 & May 2024 & 26\% & 78 & 1120 & 3958 \\
w\_avg\_1y\_SM\_combined\_past & Jun 2007 & May 2024 & 9\% & 80 & 1193 & 5002 \\
w\_avg\_1y\_TEMP\_crop & Jun 2007 & May 2024 & 25\% & 78 & 1139 & 4046 \\
w\_avg\_1y\_TEMP\_past & Jun 2007 & May 2024 & 5\% & 80 & 1228 & 5255 \\
w\_avg\_2y\_FPAR\_crop & Jun 2007 & May 2024 & 25\% & 78 & 1139 & 4046 \\
w\_avg\_2y\_FPAR\_past & Jun 2007 & May 2024 & 5\% & 80 & 1228 & 5255 \\
h\_food\_price\_index & Jul 2008 & Apr 2024 & 32\% & 38 & 573 & 3353 \\
c\_food\_price\_index & Jul 2008 & Apr 2024 & 32\% & 38 & 573 & 3353 \\
l\_food\_price\_index & Jul 2008 & Apr 2024 & 32\% & 38 & 573 & 3353 \\
o\_food\_price\_index & Jul 2008 & Apr 2024 & 32\% & 38 & 573 & 3353 \\
h\_food\_price\_index\_moving\_average & Jul 2008 & Apr 2024 & 32\% & 38 & 573 & 3353 \\
h\_food\_price\_index\_moving\_average\_deviation & Jul 2008 & Apr 2024 & 32\% & 38 & 573 & 3353 \\
h\_food\_price\_index\_z\_score & Jul 2008 & Apr 2024 & 32\% & 38 & 573 & 3353 \\
c\_food\_price\_index\_moving\_average & Jul 2008 & Apr 2024 & 32\% & 38 & 573 & 3353 \\
c\_food\_price\_index\_moving\_average\_deviation & Jul 2008 & Apr 2024 & 32\% & 38 & 573 & 3353 \\
c\_food\_price\_index\_z\_score & Jul 2008 & Apr 2024 & 32\% & 38 & 573 & 3353 \\
l\_food\_price\_index\_moving\_average & Jul 2008 & Apr 2024 & 32\% & 38 & 573 & 3353 \\
l\_food\_price\_index\_moving\_average\_deviation & Jul 2008 & Apr 2024 & 32\% & 38 & 573 & 3353 \\
l\_food\_price\_index\_z\_score & Jul 2008 & Apr 2024 & 32\% & 38 & 573 & 3353 \\
o\_food\_price\_index\_moving\_average & Jul 2008 & Apr 2024 & 32\% & 38 & 573 & 3353 \\
o\_food\_price\_index\_moving\_average\_deviation & Jul 2008 & Apr 2024 & 32\% & 38 & 573 & 3353 \\
o\_food\_price\_index\_z\_score & Jul 2008 & Apr 2024 & 32\% & 38 & 573 & 3353 \\
gdp\_growth\_filled\_lag2y & Jun 2007 & May 2024 & 0\% & 79 & 1256 & 5508 \\
INFORM\_Risk\_Index\_lag1 & Jan 2017 & May 2024 & 41\% & 72 & 1138 & 5507 \\
ucdp\_deaths & Jun 2007 & May 2024 & 0\% & 77 & 1192 & 5508 \\
ucdp\_events & Jun 2007 & May 2024 & 0\% & 77 & 1192 & 5508 \\
acled\_fatalities & Jul 2008 & May 2024 & 11\% & 75 & 1197 & 5508 \\
acled\_events & Jul 2008 & May 2024 & 11\% & 75 & 1197 & 5508 \\
idps & Jan 2019 & May 2024 & 56\% & 61 & 869 & 5477 \\
idps\_3m\_sum & Jan 2019 & May 2024 & 56\% & 61 & 869 & 5477 \\
idps\_6m\_sum & Jan 2019 & May 2024 & 56\% & 61 & 869 & 5477 \\
idps\_9m\_sum & Jan 2019 & May 2024 & 56\% & 61 & 869 & 5477 \\
Conflict Stock Displacement & Jan 2008 & May 2024 & 0\% & 80 & 1264 & 5508 \\
Conflict Internal Displacements & Jan 2008 & May 2024 & 0\% & 80 & 1264 & 5508 \\
Disaster Internal Displacements & Jan 2008 & May 2024 & 0\% & 80 & 1264 & 5508 \\
Disaster Stock Displacement & Jan 2008 & May 2024 & 0\% & 80 & 1264 & 5508 \\
Mixed Disasters & Jun 2007 & May 2024 & 0\% & 80 & 1264 & 5508 \\
Weighted Total Deaths & Jun 2007 & May 2024 & 0\% & 80 & 1264 & 5508 \\
Drought & Jun 2007 & May 2024 & 0\% & 80 & 1264 & 5508 \\
Flood & Jun 2007 & May 2024 & 0\% & 80 & 1264 & 5508 \\
Storm & Jun 2007 & May 2024 & 0\% & 80 & 1264 & 5508 \\
Mixed Disasters\_3m\_sum & Jun 2007 & May 2024 & 0\% & 80 & 1264 & 5508 \\
Mixed Disasters\_6m\_sum & Jun 2007 & May 2024 & 0\% & 80 & 1264 & 5508 \\
Mixed Disasters\_9m\_sum & Jun 2007 & May 2024 & 0\% & 80 & 1264 & 5508 \\
Weighted Total Deaths\_3m\_sum & Jun 2007 & May 2024 & 0\% & 80 & 1264 & 5508 \\
Weighted Total Deaths\_6m\_sum & Jun 2007 & May 2024 & 0\% & 80 & 1264 & 5508 \\
Weighted Total Deaths\_9m\_sum & Jun 2007 & May 2024 & 0\% & 80 & 1264 & 5508 \\
Drought\_3m\_sum & Jun 2007 & May 2024 & 0\% & 80 & 1264 & 5508 \\
Drought\_6m\_sum & Jun 2007 & May 2024 & 0\% & 80 & 1264 & 5508 \\
Drought\_9m\_sum & Jun 2007 & May 2024 & 0\% & 80 & 1264 & 5508 \\
Flood\_3m\_sum & Jun 2007 & May 2024 & 0\% & 80 & 1264 & 5508 \\
Flood\_6m\_sum & Jun 2007 & May 2024 & 0\% & 80 & 1264 & 5508 \\
Flood\_9m\_sum & Jun 2007 & May 2024 & 0\% & 80 & 1264 & 5508 \\
Storm\_3m\_sum & Jun 2007 & May 2024 & 0\% & 80 & 1264 & 5508 \\
Storm\_6m\_sum & Jun 2007 & May 2024 & 0\% & 80 & 1264 & 5508 \\
Storm\_9m\_sum & Jun 2007 & May 2024 & 0\% & 80 & 1264 & 5508 \\
Mixed Disasters\_vul & Jun 2007 & May 2024 & 0\% & 80 & 1264 & 5508 \\
Weighted Total Deaths\_vul & Jun 2007 & May 2024 & 0\% & 80 & 1264 & 5508 \\
Drought\_vul & Jun 2007 & May 2024 & 0\% & 80 & 1264 & 5508 \\
Flood\_vul & Jun 2007 & May 2024 & 0\% & 80 & 1264 & 5508 \\
Storm\_vul & Jun 2007 & May 2024 & 0\% & 80 & 1264 & 5508 \\
Tot\_Pop & Jun 2007 & May 2024 & 0\% & 80 & 1264 & 5508 \\
UCentre\_Pop & Jun 2007 & May 2024 & 0\% & 80 & 1264 & 5508 \\
UCluster\_Pop & Jun 2007 & May 2024 & 0\% & 80 & 1264 & 5508 \\
Rural\_Pop & Jun 2007 & May 2024 & 0\% & 80 & 1264 & 5508 \\
Food Price Index & Jun 2007 & May 2024 & 0\% & 80 & 1264 & 5508 \\
Meat Price Index & Jun 2007 & May 2024 & 0\% & 80 & 1264 & 5508 \\
Dairy Price Index & Jun 2007 & May 2024 & 0\% & 80 & 1264 & 5508 \\
Cereals Price Index & Jun 2007 & May 2024 & 0\% & 80 & 1264 & 5508 \\
Oils Price Index & Jun 2007 & May 2024 & 0\% & 80 & 1264 & 5508 \\
Sugar Price Index & Jun 2007 & May 2024 & 0\% & 80 & 1264 & 5508 \\
\end{longtable}

\phantomsection
\subsubsection*{Cross validation datasets}
\begin{table}[ht]
\centering
\caption{Cross-validation summary statistics (median over outer folds) for the number of data points and the number of countries, as well as the positive rate (\%). Statistics are reported for datasets that exclude records with missing values (``Restricted'' dataset) and for those trained to include such records (``Full'' dataset), across different target sources and target types. }
\label{tab:cv_summary}
\resizebox{\textwidth}{!}{%
\begin{tabular}{lll rr rr rr}
\toprule
\textbf{Source} & \textbf{Target} & \textbf{Dataset} 
  & \multicolumn{2}{c}{\textbf{Data points}} 
  & \multicolumn{2}{c}{\textbf{Countries}} 
  & \multicolumn{2}{c}{\textbf{Positive rate (\%)}} \\
\cmidrule(lr){4-5} \cmidrule(lr){6-7} \cmidrule(lr){8-9}
 &  &  & Train & Test & Train & Test & Train & Test \\
\midrule
\multirow{6}{*}{FEWS NET}
  & \multirow{2}{*}{Crisis occurrence}
    & Restricted & 6{,}001  & 1{,}570 & 13 & 13 & 40 & 28 \\
  & & Full       & 52{,}269 & 2{,}273 & 15 & 14 & 10 & 21 \\
\cmidrule(l){2-9}
  & \multirow{2}{*}{Emergency occurrence}
    & Restricted & 6{,}001  & 1{,}570 & 13 & 13 &  3 &  1 \\
  & & Full       & 52{,}269 & 2{,}273 & 15 & 14 &  1 &  2 \\
\cmidrule(l){2-9}
  & \multirow{2}{*}{Crisis escalation}
    & Restricted & 6{,}001  & 1{,}570 & 13 & 13 &  4 &  5 \\
  & & Full       & 54{,}038 & 2{,}562 & 15 & 14 &  3 &  5 \\
\midrule
\multirow{6}{*}{IPC/CH}
  & \multirow{2}{*}{Crisis occurrence}
    & Restricted &  9{,}581 & 2{,}014 & 14 &  9 & 30 & 21 \\
  & & Full       & 30{,}050 & 3{,}408 & 28 & 23 & 42 & 43 \\
\cmidrule(l){2-9}
  & \multirow{2}{*}{Emergency occurrence}
    & Restricted &  9{,}581 & 2{,}014 & 14 &  9 &  3 &  1 \\
  & & Full       & 30{,}050 & 3{,}408 & 28 & 23 &  5 &  4 \\
\cmidrule(l){2-9}
  & \multirow{2}{*}{Crisis escalation}
    & Restricted &  9{,}601 & 2{,}034 & 14 &  9 &  3 &  2 \\
  & & Full       & 30{,}274 & 3{,}972 & 30 & 22 &  4 &  3 \\
\bottomrule
\end{tabular}}
\end{table}

\begin{figure}
\includegraphics[width=0.9\textwidth]{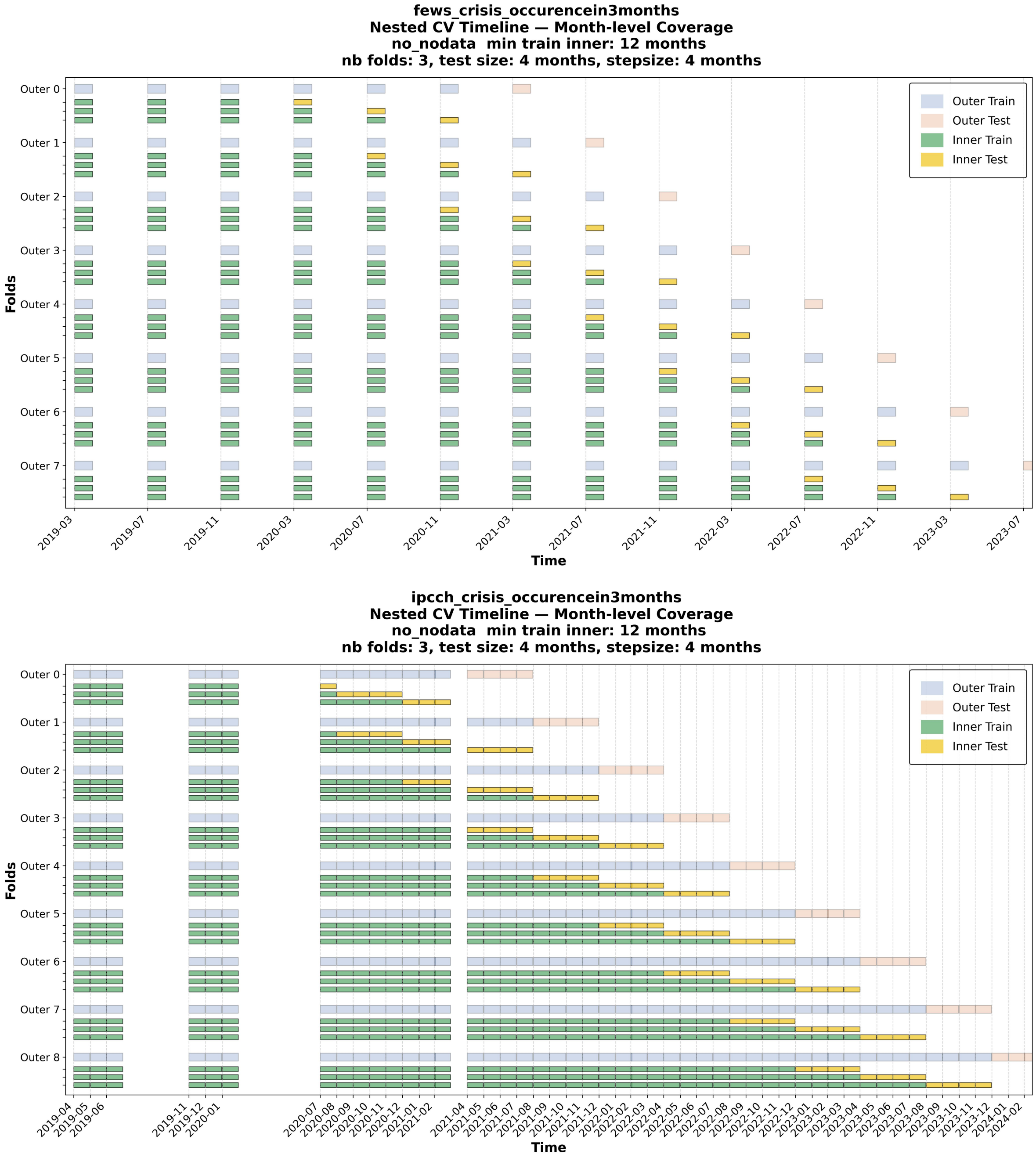}
\caption{Temporal layout of the nested cross-validation (CV) splits, at monthly resolution, showing outer and inner train/test splits. The display is for the target crisis occurrence predicted at a 3-month lead time, using FEWS NET (top) and IPC/CH (bottom) as target sources. The scheme uses 3 inner folds, a 4-month test window, and a minimum training window of 12 months. Splits are shown for the ``Restricted dataset'', which excludes records with missing target observations.}
\label{fig:cval_timeline}
\end{figure}

\phantomsection
\subsubsection*{Pipelines parameters}
\label{pipe_params}

Table~\ref{tab:hyperparam_grid} summarizes the hyperparameter search grids used during pipeline selection. For each pipeline, hyperparameters were optimized via exhaustive randomized search combined with cross-validation. All pipelines share a common feature selection step based  ``Minimum Redundancy - Maximum Relevance'' algorithm \cite{ding2005minimum} using the F-statistic for relevance and Pearson for redundancy. The number of selected features was tuned over \{20, 30, 40, 50\}. Regularization strength was tuned for logistic regression, while tree-based models were searched over varying depths and split criteria. Ensemble methods — random forest, XGBoost, and histogram-based gradient boosting — were also explored with respect to the number of estimators and learning rates. Early stopping was enabled for both XGBoost and histogram-based gradient boosting to prevent overfitting during training.

\begin{table}[ht]
\centering
\caption{Hyperparameter search grids for each pipeline.}
\label{tab:hyperparam_grid}
\begin{tabular}{llp{7cm}}
\toprule
\textbf{Pipeline} & \textbf{Hyperparameter} & \textbf{Values} \\
\midrule
\multirow{2}{*}{Logistic regression}
  & \texttt{C}       & 0.1, 1.0, 10.0 \\
  & \texttt{penalty} & l2 \\
\midrule
\multirow{3}{*}{Decision tree}
  & \texttt{max\_depth}         & None, 10, 20, 30 \\
  & \texttt{min\_samples\_split} & 2, 5, 10 \\
  & \texttt{criterion}          & gini, entropy \\
\midrule
\multirow{5}{*}{Random forest}
  & \texttt{n\_estimators}      & 200, 500, 800 \\
  & \texttt{max\_depth}         & None, 10, 20, 30 \\
  & \texttt{min\_samples\_split} & 2, 5, 10 \\
  & \texttt{criterion}          & gini, entropy \\
  & \texttt{max\_features}      & sqrt, log2, 0.8 \\
\midrule
\multirow{6}{*}{XGBoost}
  & \texttt{n\_estimators}      & 300, 600, 1000 \\
  & \texttt{max\_depth}         & 3, 6, 9 \\
  & \texttt{learning\_rate}     & 0.01, 0.05, 0.1 \\
  & \texttt{subsample}          & 0.6, 0.8, 1.0 \\
  & \texttt{colsample\_bytree}  & 0.6, 0.8, 1.0 \\
  & \texttt{early\_stopping}    & True \\
\midrule
\multirow{5}{*}{Hist.\ gradient boosting}
  & \texttt{learning\_rate}     & 0.03, 0.05 \\
  & \texttt{max\_iter}          & 300, 600 \\
  & \texttt{max\_depth}         & None, 5 \\
  & \texttt{min\_samples\_leaf} & 20, 50 \\
  & \texttt{early\_stopping}    & True \\
\midrule
All pipelines & \texttt{selector\_\_k} & 20, 30, 40, 50 \\
\bottomrule
\end{tabular}
\end{table}

\subsubsection*{Performance}

\begin{figure}[h!]
\centering
\includegraphics[width=0.9\textwidth]{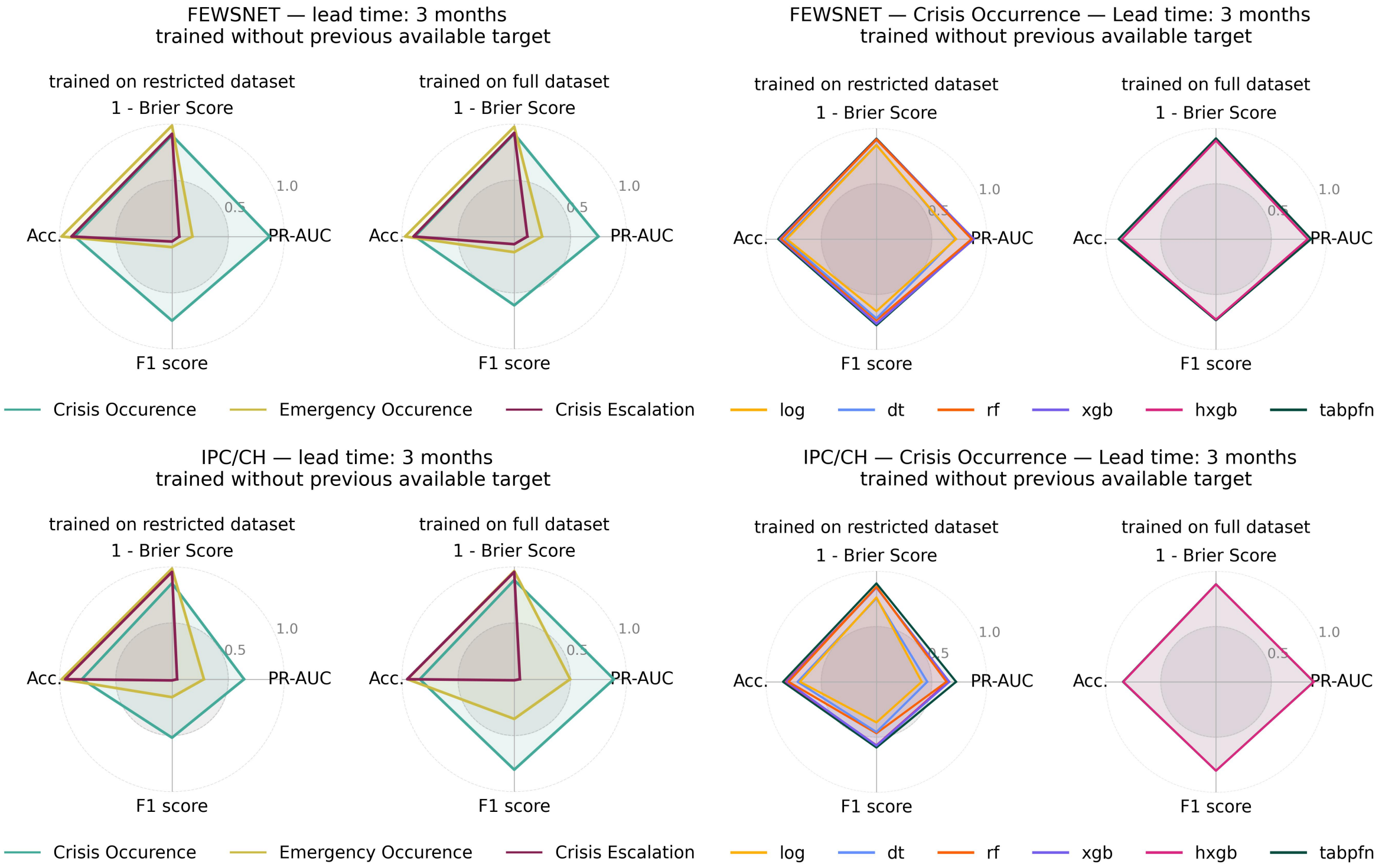}
\caption{Average metric over all test outer folds for pipelines  (log, rf, dt, xgb, hxgb, tabpfn) trained with a three‑month lead time and not counting the last available target in the features. Metrics are presented (left) averaged over all tested pipelines and for all target variables, and (right) detailed for all tested pipelines predicting crisis occurrence. Results are reported separately for pipelines predicting (top) FEWS NET and (bottom) IPC‑CH target variables and disaggregated by pipelines trained without records containing “no‑data” values (``Restricted dataset'') and for those trained with such records (full dataset).}
\label{fig:perf_radar}
\end{figure}

\begin{figure}[h!]
\includegraphics[width=0.95\textwidth]{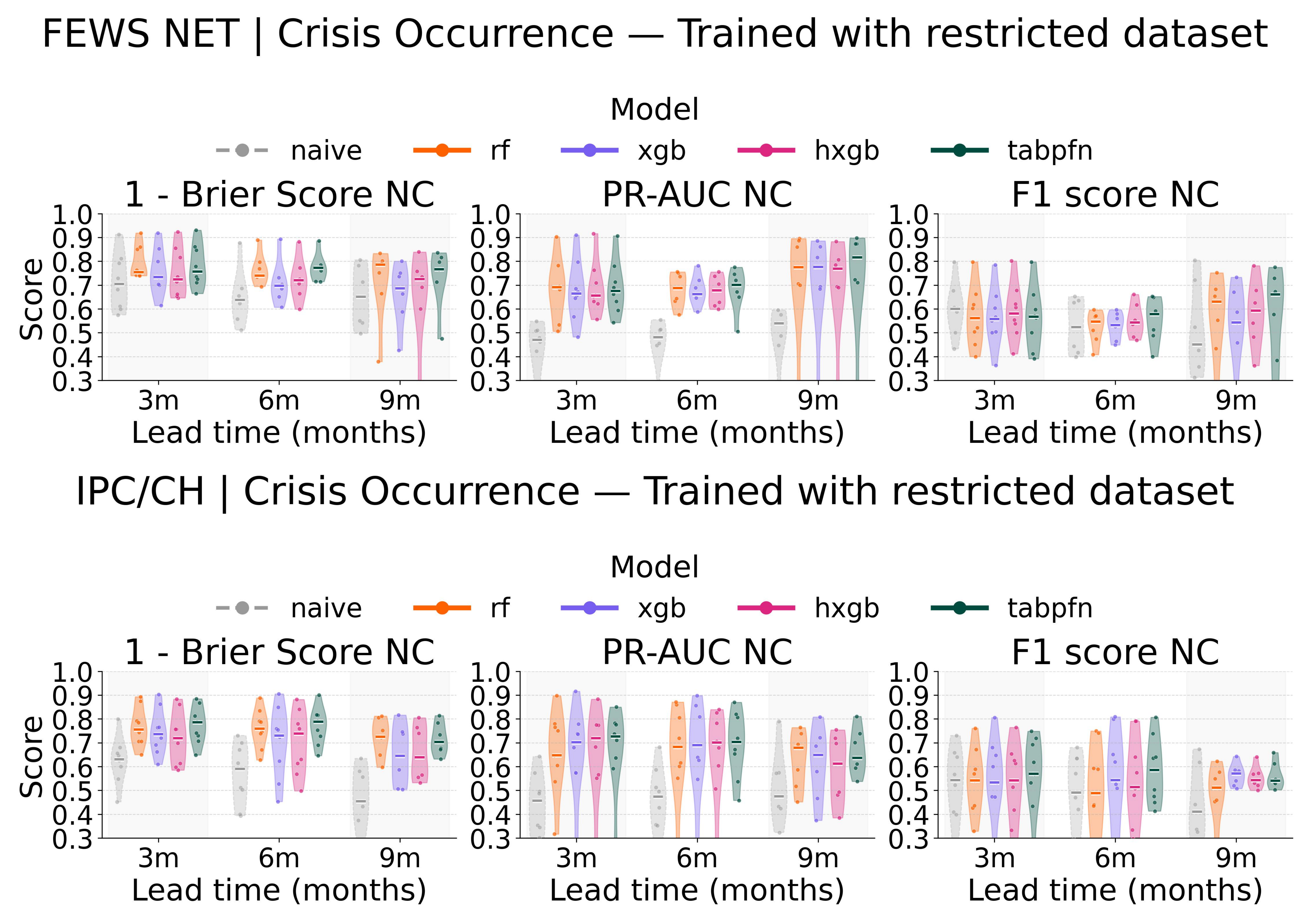}
\caption{Cross-validated pipeline performance across target sources and lead times. Outlying points, violin distributions, and median horizontal bars depict per-fold test metrics for folds with a positive rate $>10\%$, restricted to time-varying administrative areas. Random Forest (rf), XGBoost (xgb), HistXGBoost (hxgb), and TabPFN (tabpfn) are evaluated on the missing-data-free dataset (``Restricted dataset''). Top panels display FEWS NET targets; bottom panels display IPC/CH. The complement of the Brier Score is shown ($1 - \text{Brier Score}$), so that higher values indicate superior performance across all metrics.}
\label{fig:perf_violin_restricted}
\end{figure}

\begin{figure}[h!]
\centering
\includegraphics[width=16.5cm,height=7.5cm]{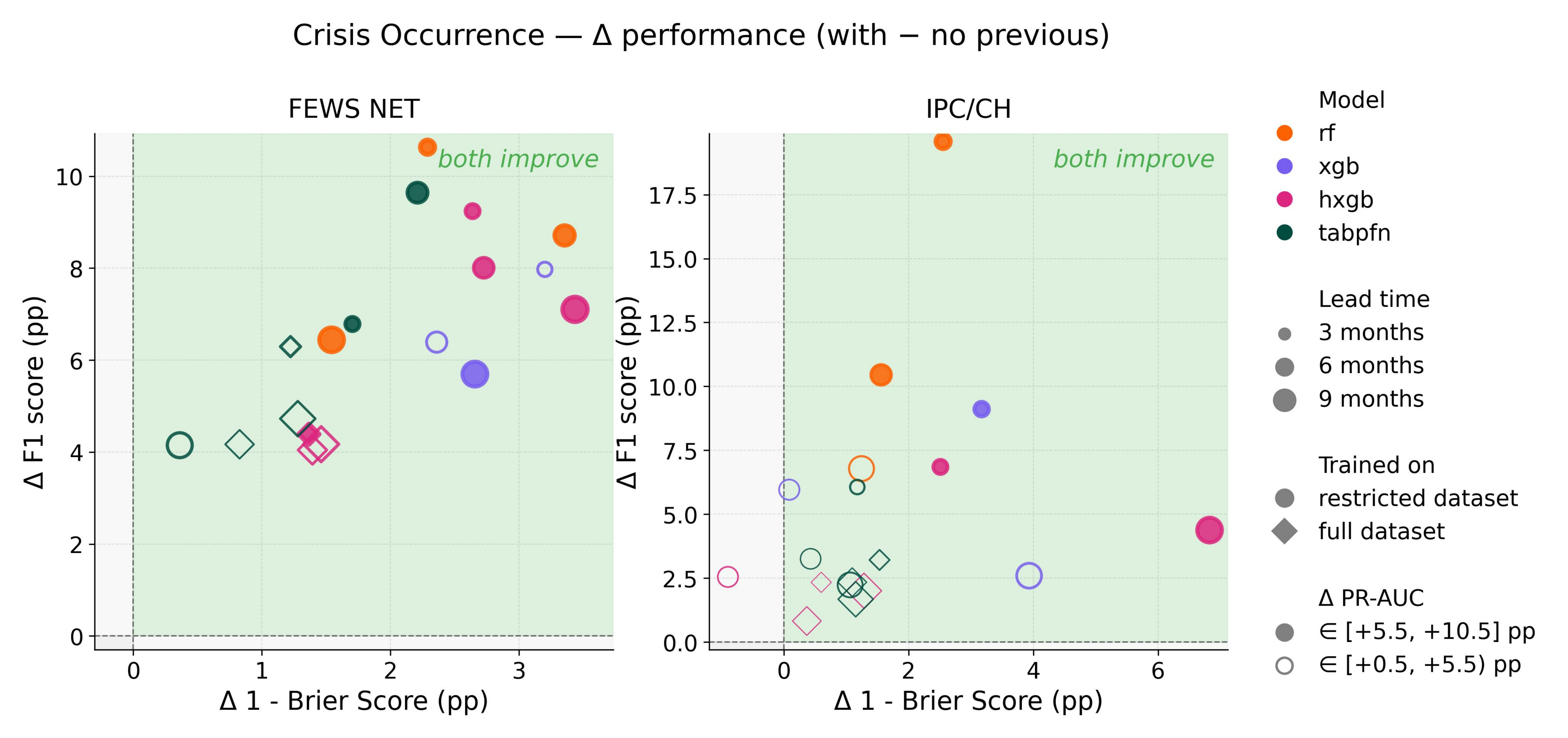}
\caption{Performance improvement from incorporating historical target information. The figure shows the difference in F1-score, 1-Brier Score (complement to Brier Score for interpretability), and PR-AUC when pipelines are trained with versus without the previous last available crisis occurrence target variable as a feature, stratified by lead time (3, 6, 9 months) and dataset completeness (``Restricted/Full dataset''). Positive values indicate performance gains from using historical information.}
\label{fig:includeprevious}
\end{figure}

\begin{table}[h!]
\centering
\caption{Mean ($\pm$ std) performance metrics across outer test folds for TabPFN (tabpfn), 
HistXGBoost(hxgb) and the naive baseline, for crisis occurrence at 
3-month lead time. Metrics are reported on all records (All) and for non-constant administrative areas (NC). AUC: PR-AUC; BS: Brier Score; 
ACC: Accuracy; F1: F1-Score.}
\label{tab:appendix_metrics}
\resizebox{\textwidth}{!}{%
\begin{tabular}{ll rrrr rrrr}
\toprule
& & \multicolumn{4}{c}{\textbf{All}} & \multicolumn{4}{c}{\textbf{Non-Constant (NC)}} \\
\cmidrule(lr){3-6} \cmidrule(lr){7-10}
\textbf{Source} & \textbf{Pipeline} & \textbf{AUC} & \textbf{BS} & \textbf{ACC} & \textbf{F1} 
                                & \textbf{AUC} & \textbf{BS} & \textbf{ACC} & \textbf{F1} \\
\midrule
\multirow{3}{*}{IPC/CH}
  & tabpfn & $.91_{\pm .04}$ & $.10_{\pm .03}$ & $.87_{\pm .06}$ & $.84_{\pm .06}$ 
           & $.87_{\pm .08}$ & $.16_{\pm .05}$ & $.78_{\pm .07}$ & $.79_{\pm .07}$ \\
  & hxgb   & $.89_{\pm .05}$ & $.11_{\pm .04}$ & $.85_{\pm .05}$ & $.83_{\pm .04}$ 
           & $.84_{\pm .09}$ & $.18_{\pm .06}$ & $.76_{\pm .07}$ & $.77_{\pm .07}$ \\
  & naive  & $.42_{\pm .10}$ & $.16_{\pm .07}$ & $.84_{\pm .07}$ & $.81_{\pm .06}$ 
           & $.53_{\pm .09}$ & $.26_{\pm .07}$ & $.74_{\pm .07}$ & $.74_{\pm .08}$ \\
\midrule
\multirow{3}{*}{FEWS NET}
  & tabpfn & $.89_{\pm .03}$ & $.08_{\pm .02}$ & $.90_{\pm .03}$ & $.80_{\pm .05}$ 
           & $.84_{\pm .06}$ & $.14_{\pm .04}$ & $.81_{\pm .05}$ & $.72_{\pm .08}$ \\
  & hxgb   & $.88_{\pm .04}$ & $.10_{\pm .03}$ & $.87_{\pm .05}$ & $.77_{\pm .05}$ 
           & $.83_{\pm .06}$ & $.17_{\pm .04}$ & $.77_{\pm .06}$ & $.72_{\pm .07}$ \\
  & naive  & $.26_{\pm .05}$ & $.10_{\pm .03}$ & $.90_{\pm .03}$ & $.81_{\pm .05}$ 
           & $.39_{\pm .09}$ & $.20_{\pm .06}$ & $.80_{\pm .06}$ & $.74_{\pm .06}$ \\
\bottomrule
\end{tabular}}
\end{table}

\begin{figure}
\includegraphics[width=0.95\textwidth]{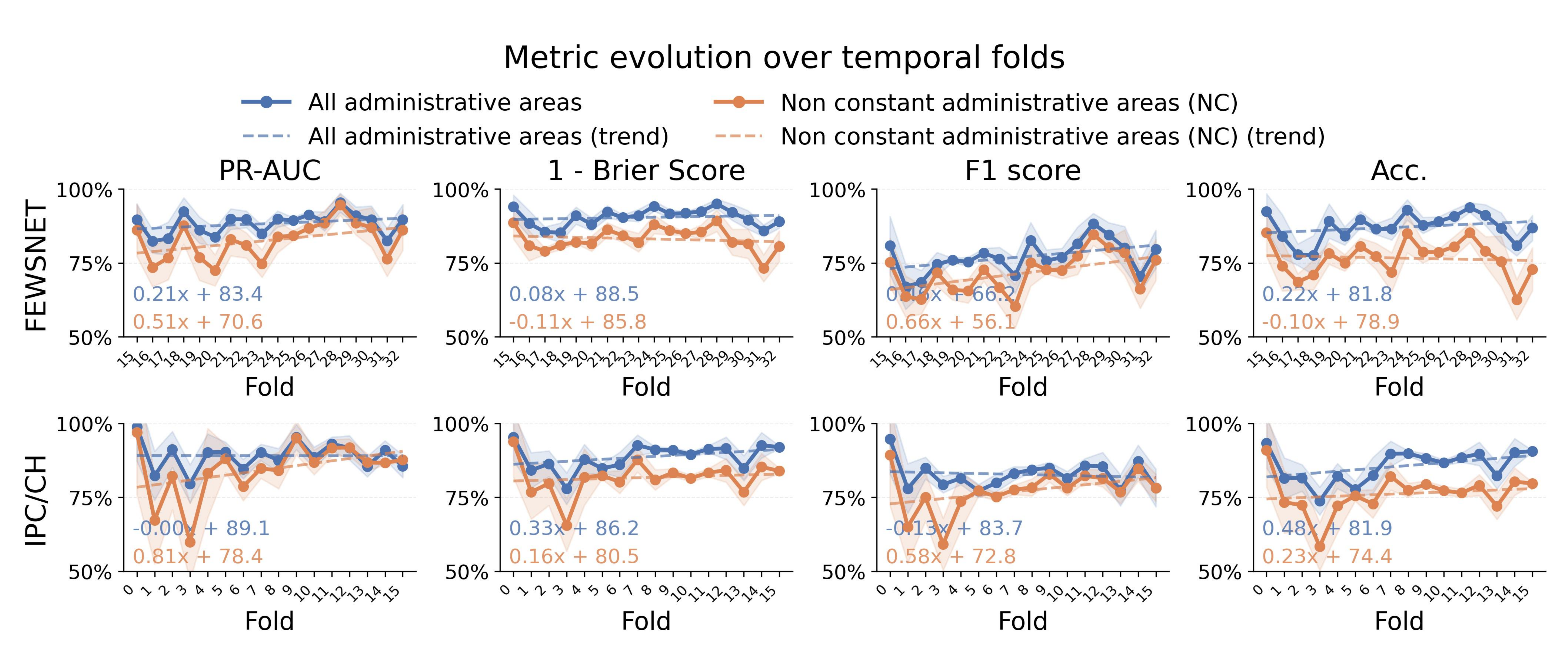}
\caption{Metrics over all test outer folds and for the best tested pipeline (HistXGBoost) for a three-month lead time and including the last available target in the features for the (top) FEWS NET and (bottom) IPC‑CH target variables on the dataset trained with records containing “no‑data” values (full dataset).}
\label{fig:temp_fold}
\end{figure}

\begin{figure}
\includegraphics[width=0.7\textwidth]{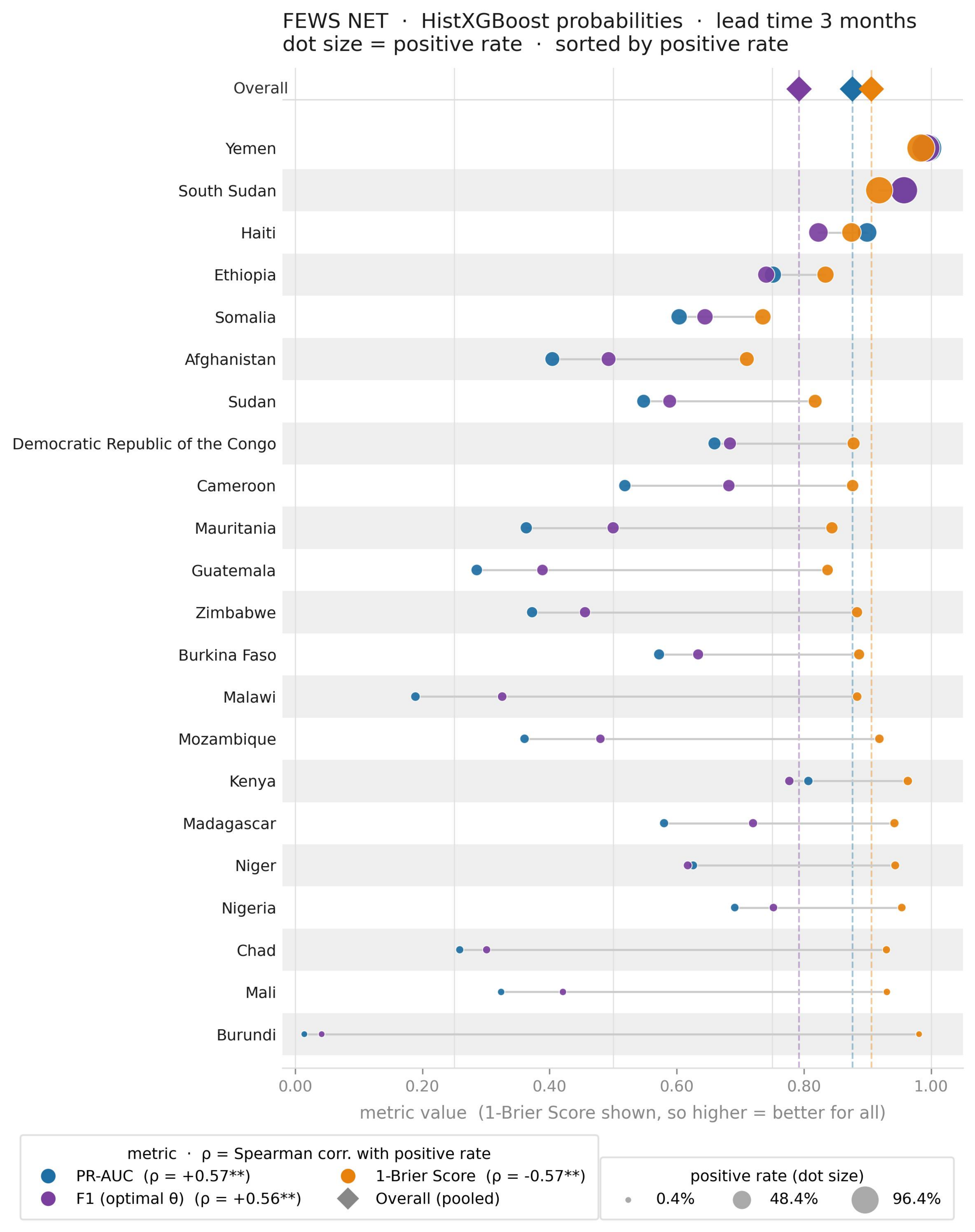}
\caption{Per-country performance of HistXGBoost trained on the FEWS NET full dataset at a 3-month lead time, sorted by PR-AUC. Each row shows PR-AUC (blue), the complement of the Brier Score (orange), and the F1-Score at the optimal threshold (purple) for a given country. Diamond markers and vertical lines indicate overall pooled performances. Dot size is proportional to the country-level positive rate.  Monotonic association between the country-level positive rate and each performance metric is assessed by Spearman rank correlation ($\rho$), reported in the legend body (significance levels: * p < 0.05, ** p < 0.01, *** p < 0.001).}
\label{fig:perf_country_fews}
\end{figure}

\begin{figure}
\includegraphics[width=0.9\textwidth]{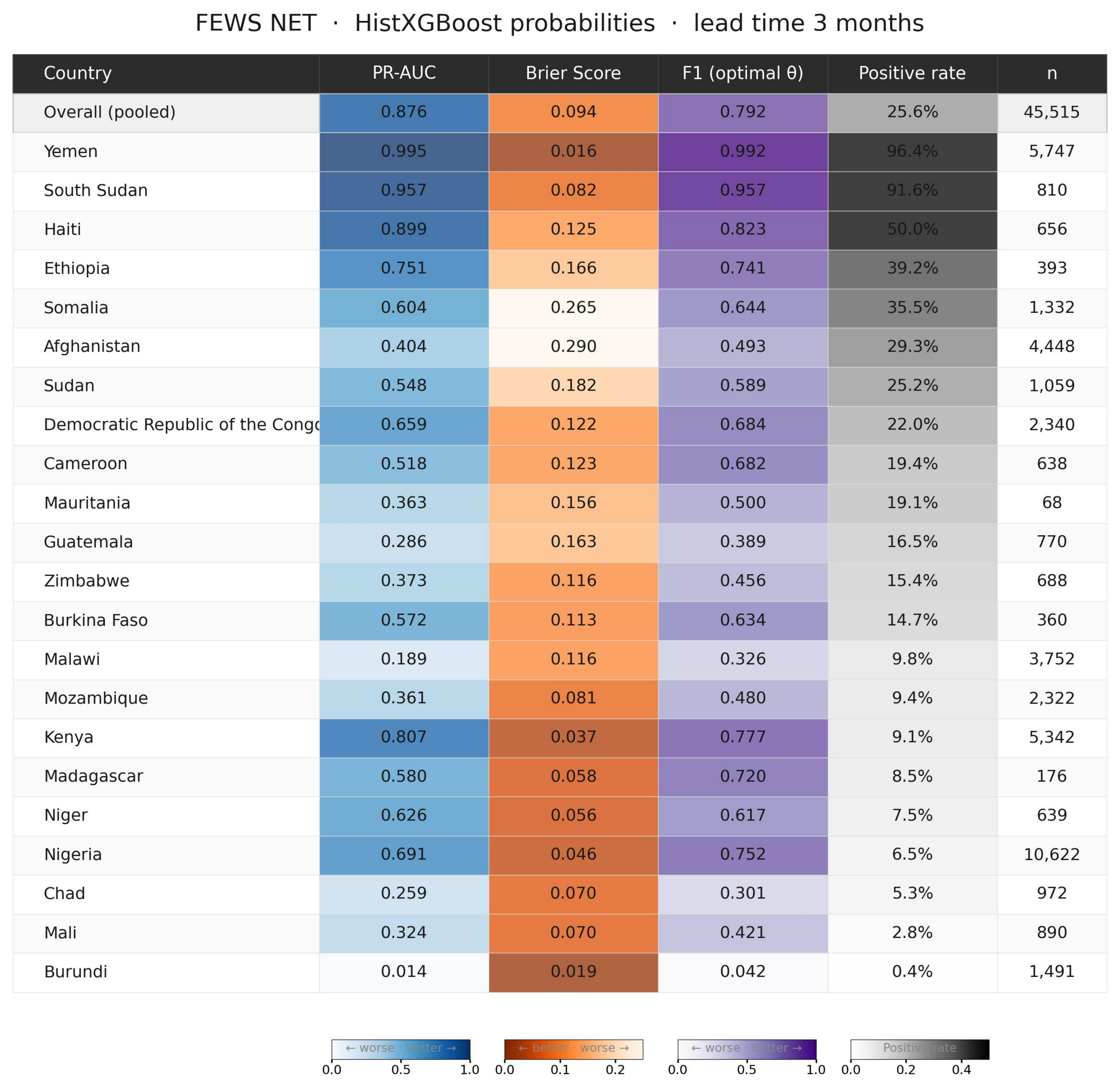}
\caption{Performance metrics per country for HistXGBoost trained on the FEWS NET full dataset at a 3-month lead time, sorted by PR-AUC. Metrics are computed on pooled test set predictions across outer folds. PR-AUC and F1-Score (at optimal threshold $\theta$) follow a higher-is-better convention; Brier Score follows a lower-is-better convention. Positive rate reflects the proportion of crisis observations per country. $n$ denotes the total number of test samples per country.}
\label{fig:perf_tab_country_fews}
\end{figure}

\begin{figure}
\includegraphics[width=0.7\textwidth]{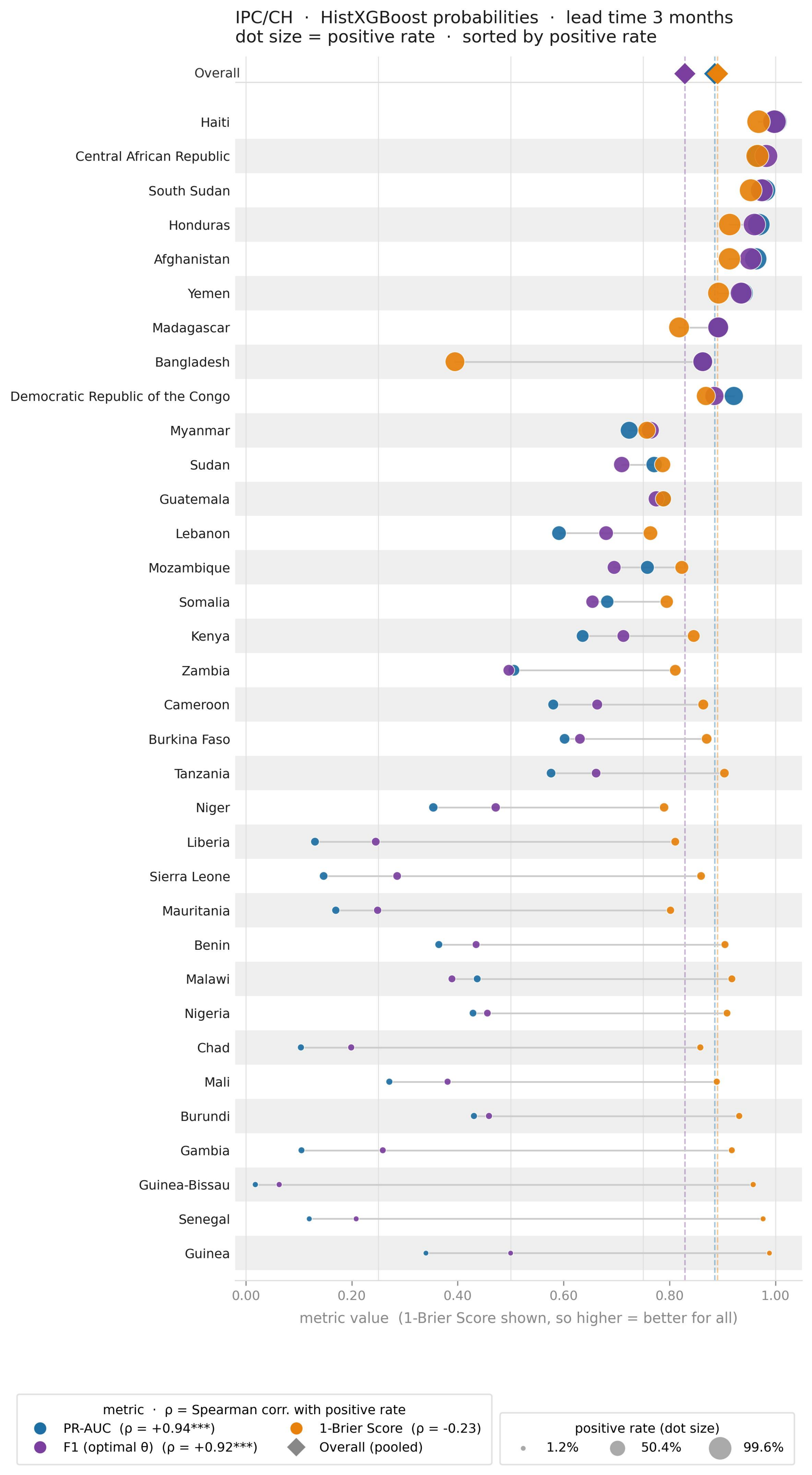}
\caption{Per-country performance of HistXGBoost trained on the IPC/CH full dataset at a 3-month lead time, sorted by PR-AUC. Each row shows PR-AUC (blue), complement to Brier Score (orange), and F1-Score at optimal threshold (purple) for a given country. Diamond markers and vertical lines indicate overall pooled performances. Dot size is proportional to the country-level positive rate.  Monotonic association between the country-level positive rate and each performance metric is assessed by Spearman rank correlation ($\rho$), reported in the legend body (significance levels: * p < 0.05, ** p < 0.01, *** p < 0.001).}
\label{fig:perf_country_ipcch}
\end{figure}

\begin{figure}
\includegraphics[width=0.9\textwidth]{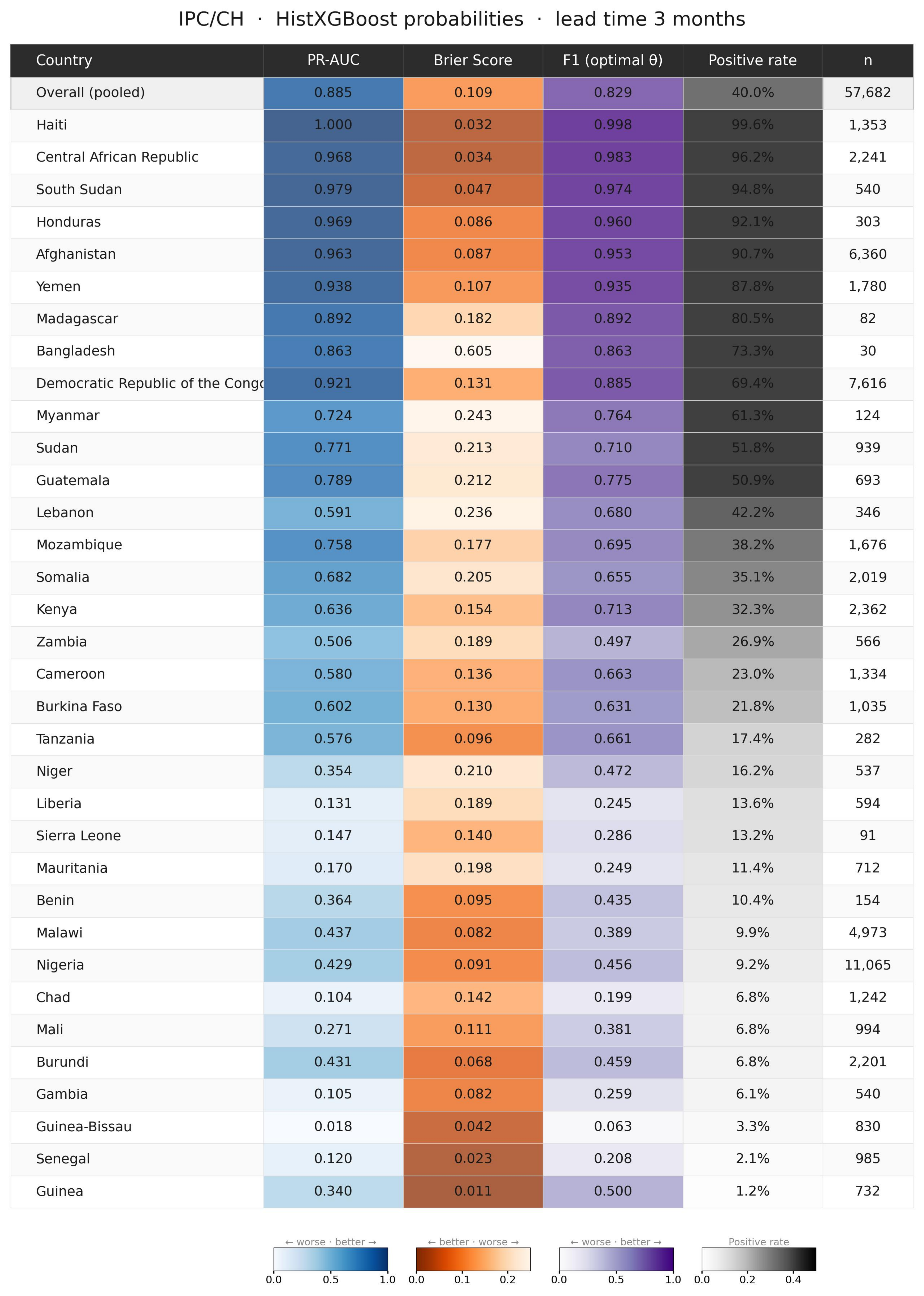}
\caption{Performance metrics per country for HistXGBoost trained on the IPC/CH full dataset at a 3-month lead time, sorted by PR-AUC. Metrics are computed on pooled test set predictions across outer folds. PR-AUC and F1-Score (at optimal threshold $\theta$) follow a higher-is-better convention; Brier Score follows a lower-is-better convention. Positive rate reflects the proportion of crisis observations per country. $n$ denotes the total number of test samples per country.}
\label{fig:perf_tab_country_ipcch}
\end{figure}

\newpage

\subsubsection*{Food crisis drivers with SHAP}

\begin{figure}[h!]
\includegraphics[width=0.9\textwidth]{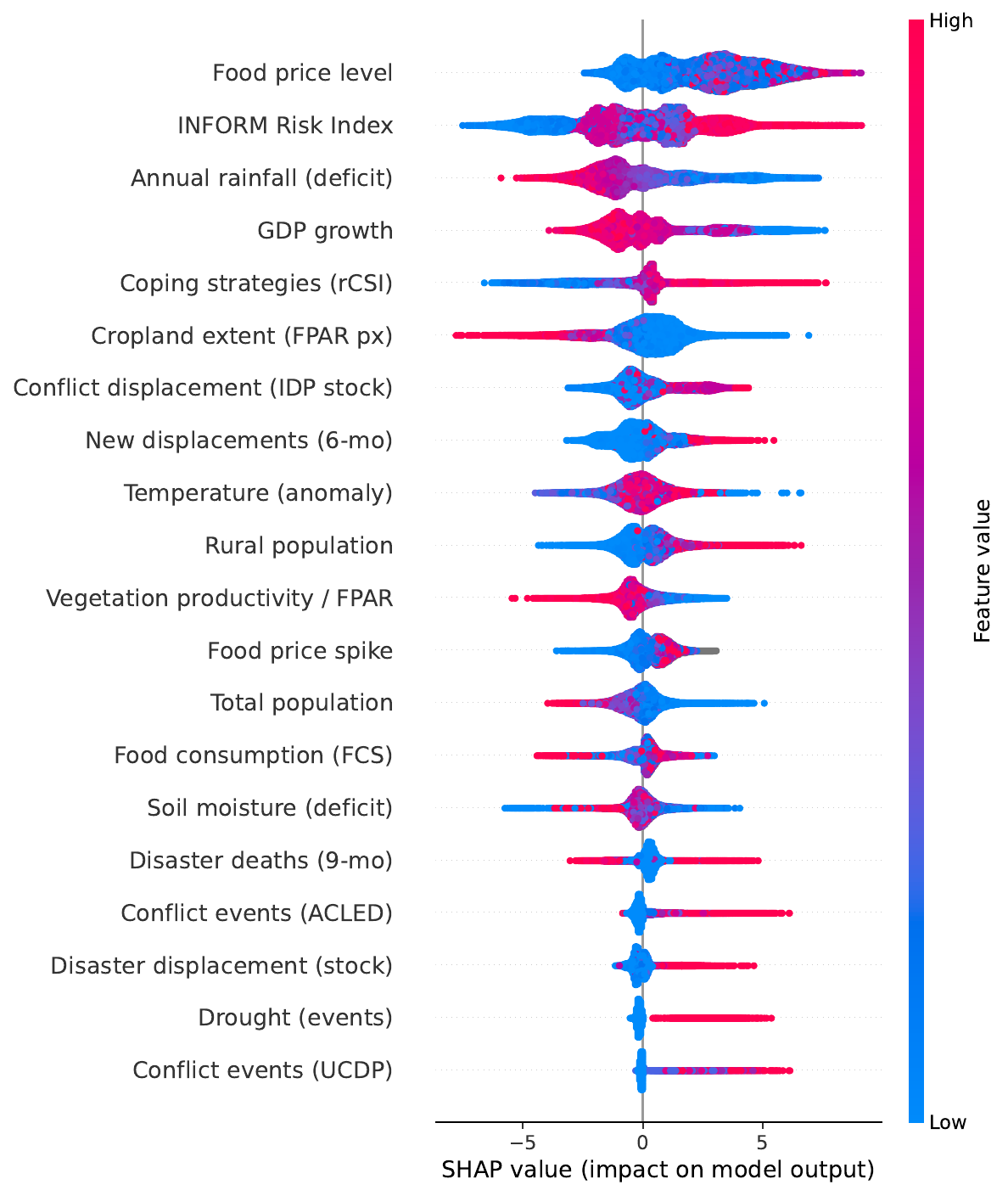}
\caption{Summary plot showing the unified SHAP values from FEWS NET and IPC/CH models at a 3-month lead time, computed using scaled values weighted by average precision scores.}
\label{fig:shap_summary}
\end{figure}

\begin{figure}[h!]
\includegraphics[width=0.9\textwidth]{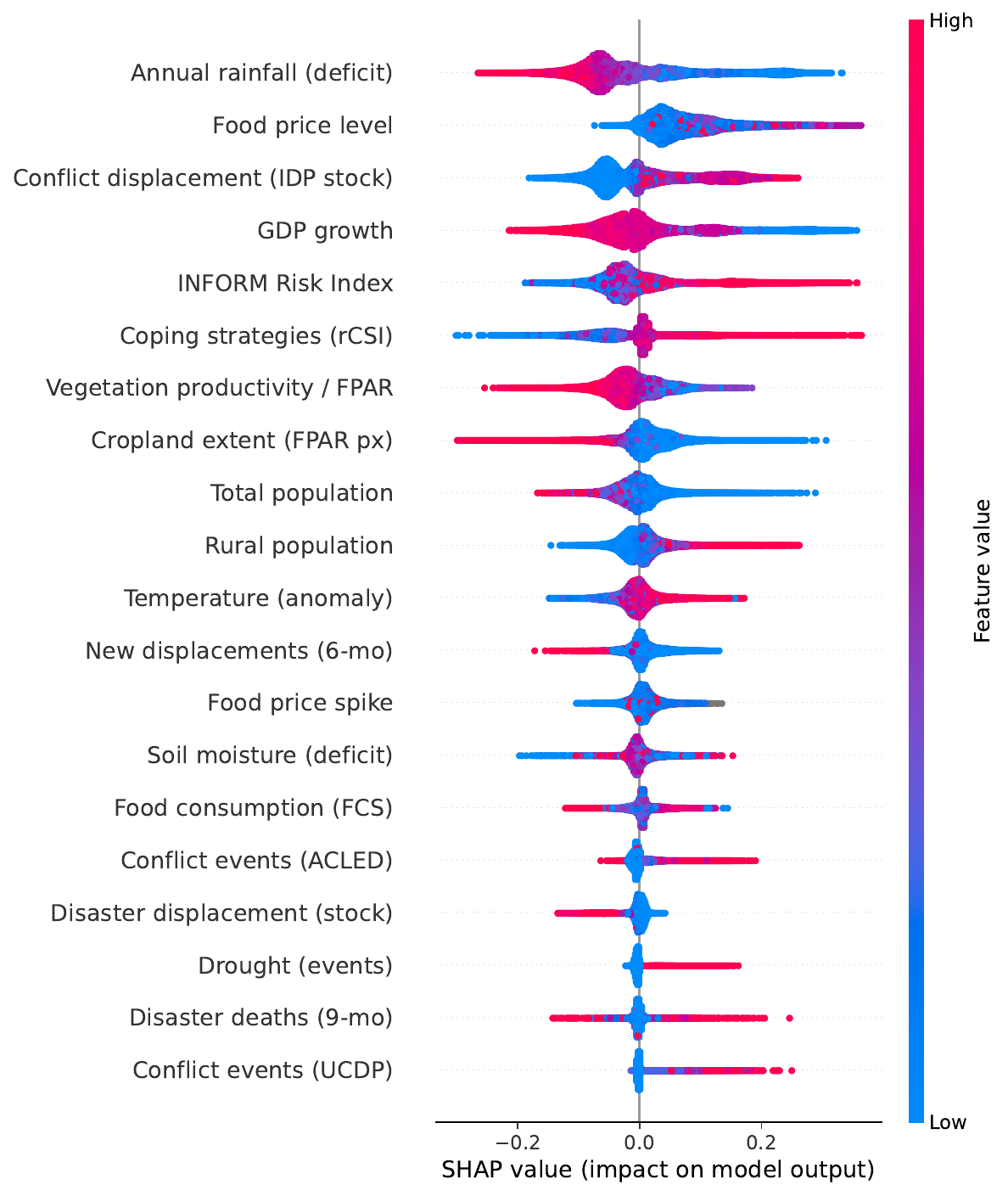}
\caption{Summary plot showing the SHAP values for the FEWS NET forecasting model at a 3-month lead time.}
\label{fig:shap_summary_fews}
\end{figure}

\begin{figure}[h!]
\includegraphics[width=0.9\textwidth]{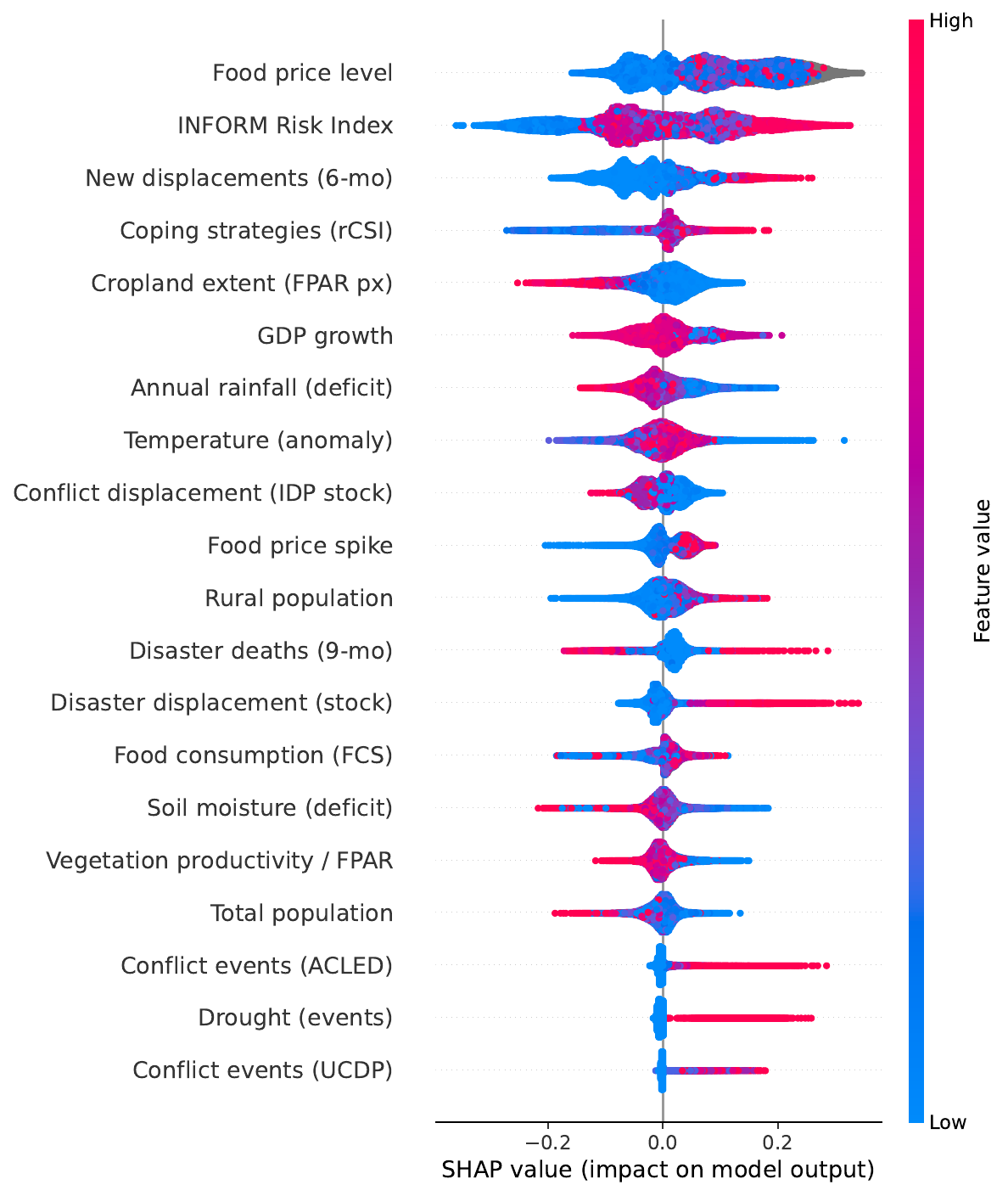}
\caption{Summary plot showing the SHAP values for the IPC/CH forecasting model at a 3-month lead time.}
\label{fig:shap_summary_ipcch}
\end{figure}

\begin{figure}[h!]
\centering
\includegraphics[width=0.9\textwidth]{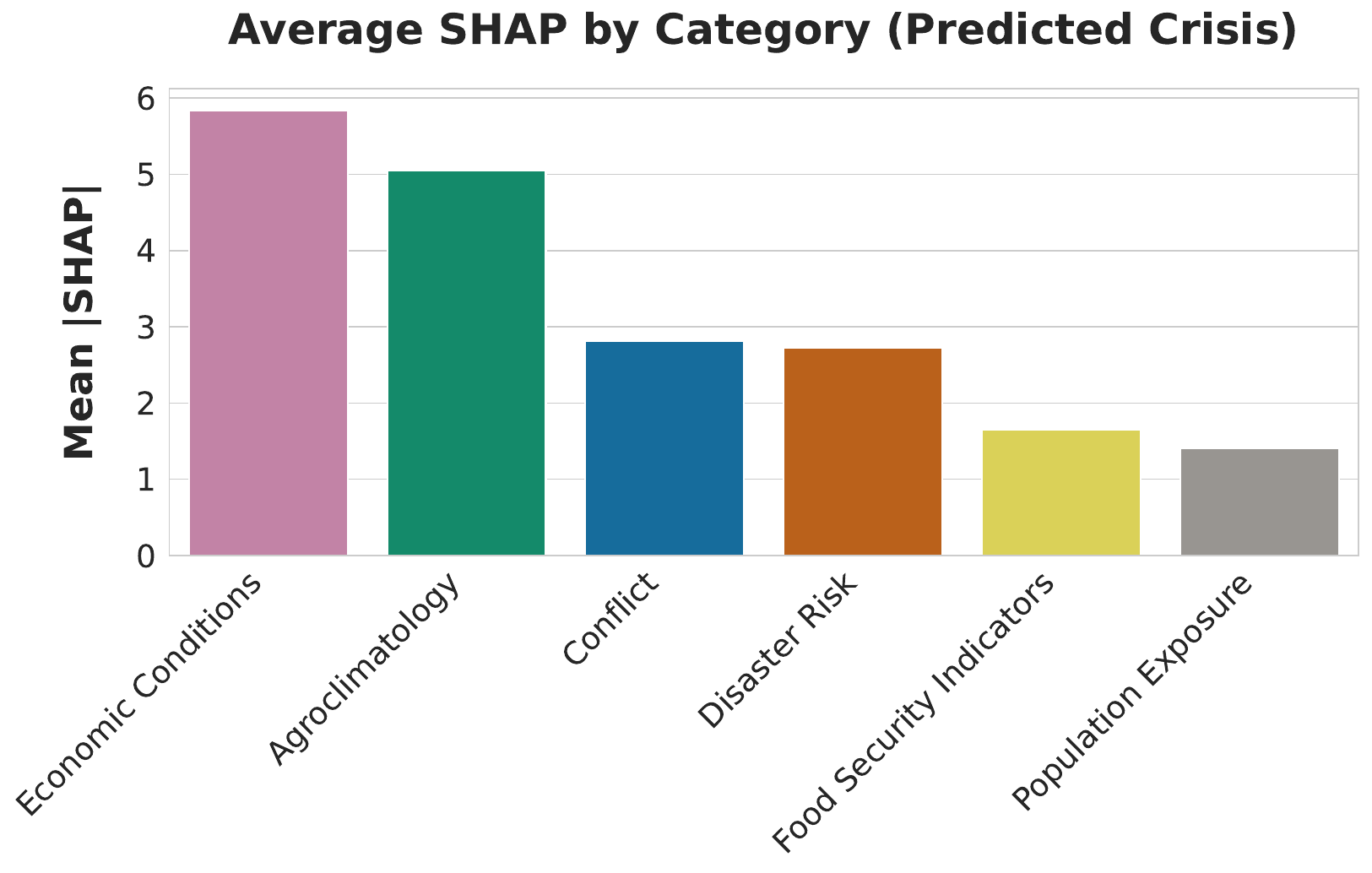}
\caption{Global feature attribution profile across thematic forecasting dimensions. Bars represent the global mean absolute unified SHAP importances for each core category, aggregated across all subnational administrative units flagged as predicted to be in a crisis state during the model evaluation period. Colors correspond to the regional risk taxonomies utilized throughout the subnational geospatial mapping framework.}
\label{fig:global_shap_bar}
\end{figure}

\begin{figure}[h!]
\centering
\includegraphics[width=18.5cm,height=9.5cm]{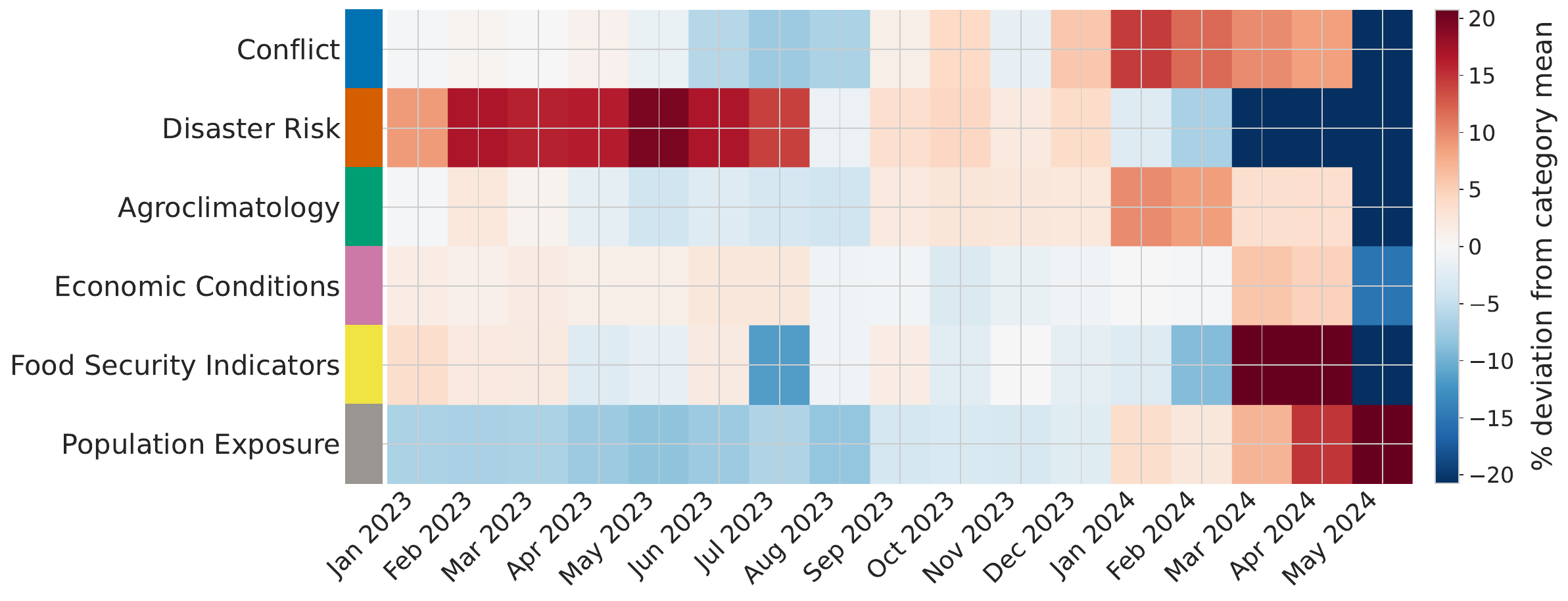}
\caption{Temporal evolution of predictive crisis signatures. The heatmap displays monthly percentage deviations of cross-sectional mean absolute unified SHAP importances relative to each dimension’s global temporal mean, tracking structural shifts in feature reliance over time. Values represent relative changes in a dimension’s predictive contribution across the evaluation horizon, highlighting periods of above- or below-average feature utilization within the unified modeling framework. Positive (red) values indicate months where a dimension contributes more than its temporal baseline, while negative (blue) values indicate below-average feature importance. Categories are ordered by overall mean absolute unified SHAP importance across the full evaluation period.}
\label{fig:shap_temporal}
\end{figure}

\begin{figure}[h!]
\centering
\includegraphics[width=18.5cm]{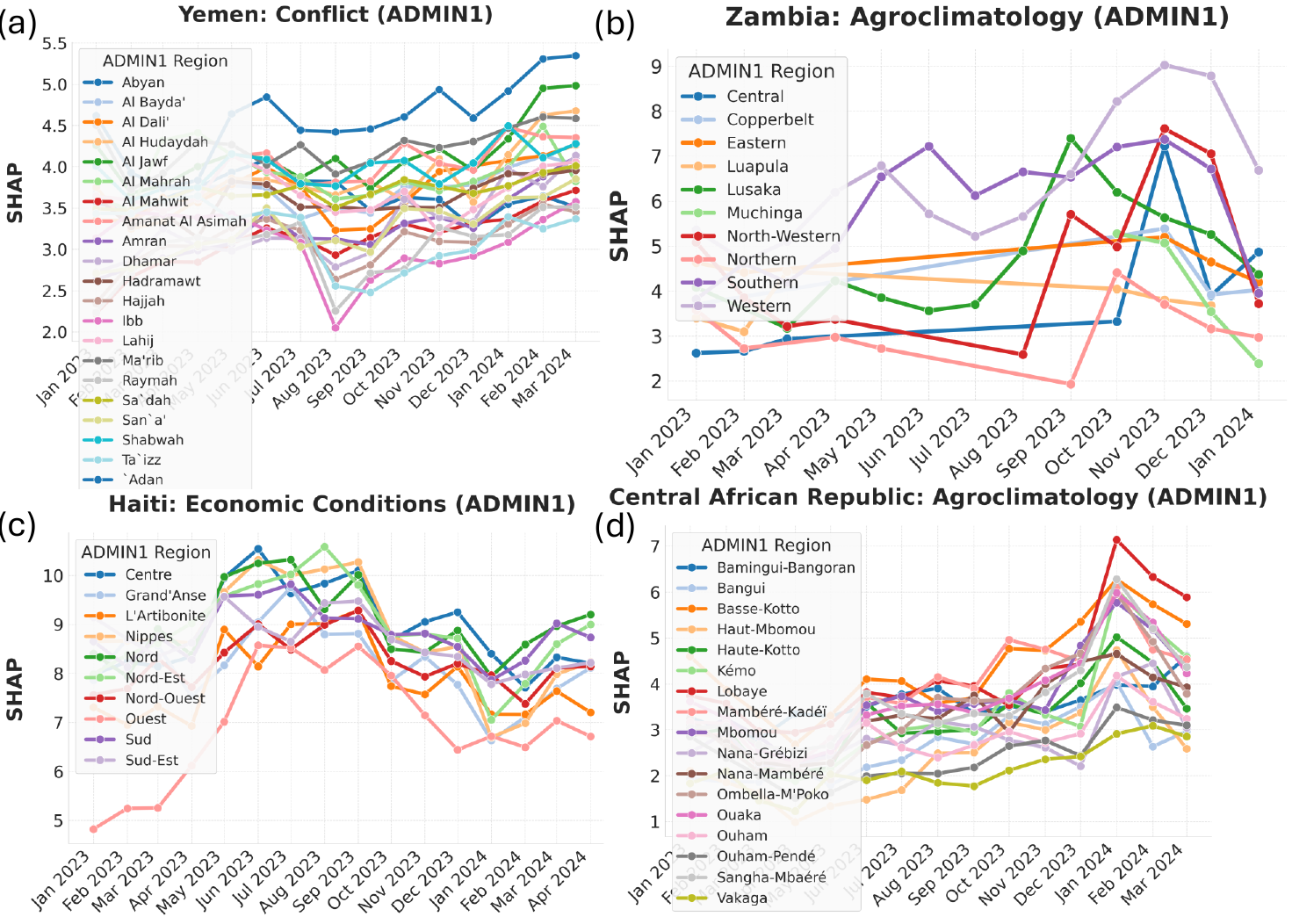}
\caption{Admin-1 aggregated attribution trajectories across selected food crisis contexts. Panels illustrate the temporal variations of regional mean absolute unified SHAP importances averaged at the Admin-1 level across different thematic forecasting categories: 
a, Conflict attribution profiles across administrative regions in Yemen. 
b, Agroclimatology attribution profiles across administrative regions in Zambia. 
c, Economic Conditions attribution profiles across administrative regions in Haiti. 
d, Agroclimatology attribution profiles across administrative regions in the Central African Republic.}
\label{fig:country_trajectories2}
\end{figure}

\begin{figure}[h!]
\centering
\includegraphics[width=18.5cm]{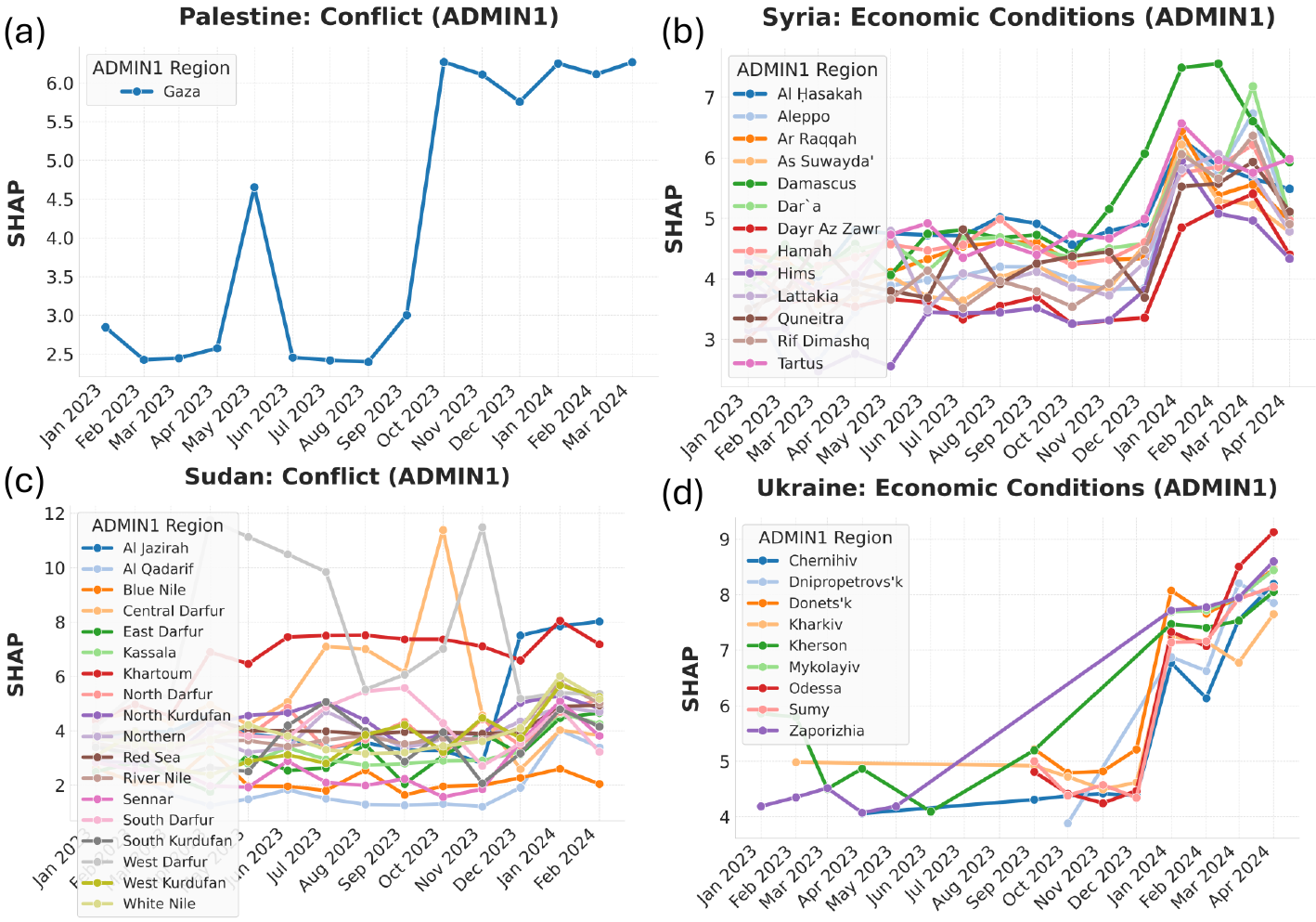}
\caption{Admin-1 aggregated attribution trajectories across selected food crisis contexts. Panels illustrate the temporal variations of regional mean absolute unified SHAP importances averaged at the Admin-1 level across different thematic forecasting categories: 
a, Conflict predictive signatures across administrative regions in Palestine. 
b, Economic Conditions predictive signatures across administrative regions in Syria. 
c, Conflict predictive signatures across administrative regions in Sudan. 
d, Economic Conditions predictive signatures across administrative regions in Ukraine.}
\label{fig:country_trajectories1}
\end{figure}

\begin{table*}[ht]
\centering
\caption{Predictive importance (SHAP) and causal estimates (ATE)
for IPC/CH classifications.}
\label{tab:ipc_drivers}

\begin{threeparttable}

\begin{adjustbox}{width=\textwidth}
\begin{tabular}{llccccccl}
\toprule
Dimension & Feature &
Mean $|\mathrm{SHAP}|$ &
ATE &
95\% CI &
Off-support &
E-val &
Status \\
\midrule

\multicolumn{8}{l}{\textbf{Conflict}}\\
Conflict & Conflict displacement (IDP stock) & 0.88 & -- & -- & 100\% & n/a & Not identified \\
Conflict & New displacements (6-mo) & 0.82 & -- & -- & 100\% & n/a & Not identified \\
Conflict & Conflict events (ACLED) & 0.35 & 1.70 & [0.8, 2.6] & 19\% & 1.25 & Causal \\
Conflict & Conflict events (UCDP) & 0.16 & 4.88 & [3.1, 6.8] & 46\% & 1.42 & Causal \\

\addlinespace
\multicolumn{8}{l}{\textbf{Disaster Risk}}\\
Disaster Risk & INFORM Risk Index & 2.05 & -- & -- & 100\% & n/a & Not identified \\
Disaster Risk & Disaster deaths (9-mo) & 0.37 & 0.18 & [-1.8, 2.1] & 59\% & 1.07 & Null \\
Disaster Risk & Disaster displacement (stock) & 0.32 & -- & -- & 99\% & n/a & Not identified \\
Disaster Risk & Drought (events) & 0.24 & 2.36 & [-0.1, 4.8] & 79\% & 1.28 & Null \\

\addlinespace
\multicolumn{8}{l}{\textbf{Agroclimatology}}\\
Agroclimatology & Annual rainfall (deficit) & 1.63 & 1.93 & [0.6, 3.2] & 68\% & 1.28 & Causal \\
Agroclimatology & Cropland extent (FPAR px) & 1.02 & -- & -- & -- & n/a & No variation \\
Agroclimatology & Temperature (anomaly) & 0.73 & 1.43 & [0.3, 2.5] & 50\% & 1.23 & Causal \\
Agroclimatology & Vegetation productivity / FPAR & 0.69 & 0.24 & [-0.9, 1.3] & 46\% & 1.08 & Null \\
Agroclimatology & Soil moisture (deficit) & 0.41 & 2.17 & [0.9, 3.4] & 65\% & 1.32 & Causal \\

\addlinespace
\multicolumn{8}{l}{\textbf{Economic Conditions}}\\
Economic Conditions & Food price level & 2.71 & 0.43 & [-0.5, 1.4] & 89\% & 1.11 & Null \\
Economic Conditions & GDP growth & 1.39 & -- & -- & 100\% & n/a & Not identified \\
Economic Conditions & Food price spike & 0.59 & 0.96 & [0.1, 1.8] & 73\% & 1.18 & Causal \\

\addlinespace
\multicolumn{8}{l}{\textbf{Food Security \& Population Exposure}}\\
Food Security Indicators & Coping strategies (rCSI) & 1.21 & -- & -- & -- & n/a & Symptom \\
Food Security Indicators & Food consumption (FCS) & 0.47 & -- & -- & -- & n/a & Symptom \\
Population Exposure & Rural population & 0.72 & -- & -- & -- & n/a & Exposure \\
Population Exposure & Total population & 0.54 & -- & -- & -- & n/a & Exposure \\

\bottomrule
\end{tabular}
\end{adjustbox}

\end{threeparttable}
\end{table*}

\begin{table*}[ht]
\centering
\caption{Predictive importance (SHAP) and causal estimates (ATE)
for FEWS NET classifications.}
\label{tab:fews_drivers}

\begin{threeparttable}

\begin{adjustbox}{width=\textwidth}
\begin{tabular}{llccccccl}
\toprule
Dimension & Feature &
Mean $|\mathrm{SHAP}|$ &
ATE &
95\% CI &
Overlap &
E-val &
Status \\
\midrule

\multicolumn{8}{l}{\textbf{Conflict}}\\
Conflict & Conflict displacement (IDP stock) & 0.88 & -- & -- & 99\% & n/a & Not identified \\
Conflict & New displacements (6-mo) & 0.82 & -- & -- & 100\% & n/a & Not identified \\
Conflict & Conflict events (ACLED) & 0.35 & 2.14 & [1.2, 3.1] & 42\% & 1.51 & Causal \\
Conflict & Conflict events (UCDP) & 0.16 & 3.42 & [2.3, 4.7] & 59\% & 1.59 & Causal \\

\addlinespace
\multicolumn{8}{l}{\textbf{Disaster Risk}}\\
Disaster Risk & INFORM Risk Index & 2.05 & -- & -- & 100\% & n/a & Not identified \\
Disaster Risk & Disaster deaths (9-mo) & 0.37 & -- & -- & 56\% & n/a & Unstable \\
Disaster Risk & Disaster displacement (stock) & 0.32 & -- & -- & 100\% & n/a & Not identified \\
Disaster Risk & Drought (events) & 0.24 & 4.64 & [3.2, 6.0] & 77\% & 2.02 & Causal \\

\addlinespace
\multicolumn{8}{l}{\textbf{Agroclimatology}}\\
Agroclimatology & Annual rainfall (deficit) & 1.63 & 0.56 & [-0.3, 1.3] & 55\% & 1.24 & Null \\
Agroclimatology & Cropland extent (FPAR px) & 1.02 & -- & -- & -- & n/a & No variation \\
Agroclimatology & Temperature (anomaly) & 0.73 & 1.95 & [1.4, 2.5] & 57\% & 1.56 & Causal \\
Agroclimatology & Vegetation productivity / FPAR & 0.69 & -0.25 & [-0.9, 0.3] & 35\% & 1.15 & Null \\
Agroclimatology & Soil moisture (deficit) & 0.41 & 2.38 & [1.6, 3.2] & 46\% & 1.61 & Causal \\

\addlinespace
\multicolumn{8}{l}{\textbf{Economic Conditions}}\\
Economic Conditions & Food price level & 2.71 & -- & -- & 97\% & n/a & Not identified \\
Economic Conditions & GDP growth & 1.39 & -- & -- & 100\% & n/a & Not identified \\
Economic Conditions & Food price spike & 0.59 & 0.29 & [-0.4, 0.9] & 84\% & 1.15 & Null \\

\addlinespace
\multicolumn{8}{l}{\textbf{Food Security \& Population Exposure}}\\
Food Security Indicators & Coping strategies (rCSI) & 1.21 & -- & -- & -- & n/a & Symptom \\
Food Security Indicators & Food consumption (FCS) & 0.47 & -- & -- & -- & n/a & Symptom \\
Population Exposure & Rural population & 0.72 & -- & -- & -- & n/a & Exposure \\
Population Exposure & Total population & 0.54 & -- & -- & -- & n/a & Exposure \\

\bottomrule
\end{tabular}
\end{adjustbox}

\end{threeparttable}
\end{table*}

\end{document}